\documentclass[10pt,letterpaper]{article}

\usepackage[top=0.85in,left=1in,footskip=0.75in]{geometry}

\usepackage{changepage}

\usepackage[utf8]{inputenc}

\usepackage{textcomp,marvosym}

\usepackage{fixltx2e}

\usepackage{amsmath,amssymb}

\usepackage{graphicx}

\usepackage{cite}

\usepackage{nameref,hyperref}

\usepackage{microtype}
\DisableLigatures[f]{encoding = *, family = * }

\usepackage{rotating}

\raggedright
\usepackage[aboveskip=1pt,labelfont=bf,labelsep=period,justification=raggedright,singlelinecheck=off]{caption}

\makeatletter
\renewcommand{\@biblabel}[1]{\quad#1.}
\makeatother

\date{}

\newcommand{\argmin}{\operatornamewithlimits{arg\ min}}

\usepackage{siunitx}
\usepackage{acronym} 
\acrodef{LIF}{Leaky Integrate-and-Fire}
\acrodef{STDP}{Spike-Timing-Dependent Plasticity}
\acrodef{PSP}{Postsynaptic Potential}
\acrodef{SRM}{Spike Response Model}
\acrodef{vRD}{van Rossum Distance}
\acrodef{ReSuMe}{Remote Supervised learning Method}
\acrodef{SEM}{Standard Error of the Mean}

\begin{document}
\vspace*{0.35in}

\begin{flushleft}
{\Large
\textbf\newline{Encoding Spike Patterns in Multilayer Spiking Neural Networks}
}
\newline
\\
Brian Gardner\textsuperscript{*},
Ioana Sporea,
André Grüning
\\
\bf{} Department of Computing, 
University of Surrey, Guildford, Surrey, United Kingdom
\\

* E-mail: b.gardner@surrey.ac.uk
\end{flushleft}

\section*{Abstract} 
Information encoding in the nervous system is supported through the precise spike-timings of neurons; however, an understanding of the underlying processes by which such representations are formed in the first place remains unclear. Here we examine how networks of spiking neurons can learn to encode for input patterns using a fully temporal coding scheme. To this end, we introduce a learning rule for spiking networks containing hidden neurons which optimizes the likelihood of generating desired output spiking patterns. 
We show the proposed learning rule allows for a large number of accurate input-output spike pattern mappings to be learnt, which outperforms other existing learning rules for spiking neural networks: both in the number of mappings that can be learnt as well as the complexity of spike train encodings that can be utilised. The learning rule is successful even in the presence of input noise, is demonstrated to solve the linearly non-separable XOR computation and generalizes well on an example dataset.
We further present a biologically plausible implementation of backpropagated learning in multilayer spiking networks, and discuss the neural mechanisms that might underlie its function.
Our approach contributes both to a systematic understanding of how pattern encodings might take place in the nervous system, and a learning rule that displays strong technical capability.

\section*{Introduction}
The importance of neuronal spike-timing in neural and cognitive information processing has been indicated at in a variety of studies \cite{Bohte2004}. For example, in the olfactory system the precision of spike-timing has been associated with accurate odor-classifications \cite{Laurent1996,Vickers2001}, and populations of auditory neurons are known to signal input features by the relative timing of spikes \cite{DeCharms1996,Knudsen2002}. However, an understanding of how the brain learns to reliably associate specific input patterns with desired spike responses through synaptic strength modifications remains a significant challenge.

From experimental observations, \acf{STDP} is widely believed to underpin learning in the brain \cite{Caporale2008}, which can induce either long- or short-term potentiation in synapses based on local variables such as the relative timing of spikes, voltage and firing frequency \cite{Bi1998,Sjostrom2001}. This closely follows Hebb's postulate for associative learning: `Cells that fire together, wire together' \cite{Hebb1949}.
Drawing on this as inspiration, a variety of supervised learning rules have been proposed that can train either single- or multilayer networks to generate desired output spike patterns in response to spatio-temporal spiking input patterns (for a recent review, see \cite{Gutig2014}).

With respect to single-layer networks, the learning rule introduced by Ponulak et al. \cite{Ponulak2010}, \acf{ReSuMe}, is a notable example of how \ac{STDP} can be applied in teaching a neuron to generate desired output spikes; from assuming an appropriate functional dependence of synaptic weight changes on the relative timing between actual and target output spikes, for example an exponential decay \cite{Gerstner2002}, the dissimilarity between an actual and target output spike train can be minimized by an analogous method to the Widrow-Hoff learning rule. Practical advantages of \ac{ReSuMe} include being independent of the neuron model implementation, and its rapid convergence when learning to perform arbitrary input-output spike pattern associations. However, although \ac{ReSuMe} retains a high degree of biological plausibility through its incorporation of \ac{STDP}, it still remains a heuristically derived learning rule, and therefore cannot be assumed to provide optimal solutions. An alternative and more rigorous formulation of \ac{ReSuMe} using gradient descent has been proposed by Sporea \& Grüning \cite{Sporea2013}.

A further supervised rule, proposed by Pfister et al. \cite{Pfister2006}, instead takes a statistical approach that optimizes by gradient ascent the likelihood of generating desired output spike times. In this case, a probabilistic spiking neuron model is used to provide a smooth functional dependence of output spike times with respect to network parameters. From simulations, the authors demonstrated a resemblance of the learning rule to experimentally observed \ac{STDP}, and demonstrated its applicability on an example `detection' learning task where input patterns were identified by the responses of readout neurons; in previous work, we have further demonstrated the high capacity achievable with this this rule when training networks to encode multiple input patterns by specific output spike patterns \cite{Gardner2014a}. An important advantage of this method are its general applicability to a wide range of learning paradigms: from supervised \cite{Pfister2006,Brea2013} to reinforcement \cite{Urbanczik2009,Fremaux2013} learning. Furthermore, it has been shown that a unique global maximum in the log-likelihood of generating a desired target spike pattern can be found when taking gradient ascent for a single-layer network \cite{Pillow2004}. Despite this, there still exists little work that has aimed at exploring its performance when encoding for a large number input spike patterns, with a reasonable number of spikes, by the precise timing of output spikes.

Additional single-layer learning rules have been developed for spiking neural network \cite{Albers2013,Memmesheimer2014,Mohemmed2012,Xu2013,Florian2012,Gutig2006}, many of which have used either an objective error function that is minimized by gradient descent or an analog of the Widrow-Hoff learning rule. Key examples include the Tempotron \cite{Gutig2006}, which has shown a strong capability in learning binary classifications of input patterns, and the Chronotron \cite{Florian2012}, that can learn to classify a large number of input patterns into multiple categories by the precise timing of output spikes.

Comparatively, the majority of research has focused on training single- rather than multilayer networks. Existing work that has examined networks containing hidden spiking neurons include SpikeProp proposed by Bohte et al. \cite{Bohte2004}, multilayer \ac{ReSuMe} by Sporea \& Grüning \cite{Sporea2013} and the recurrent network learning rules formulated by Brea et al. \cite{Brea2013} and Rezende \& Gerstner \cite{JimenezRezende2014}. Learning rules for spiking networks have proven to be a challenge to formulate, and especially given the discontinuous nature of neuronal spike-timing. A typical solution has been to assume a linear dependence of a neuron's spike-timing on presynaptic inputs around its firing threshold, such that small changes in its input with respect to synaptic weights shifts the timing of an output spike. However, such an approach has the disadvantage of constraining the learning rate to a small value \cite{Bohte2004}. An alternative approach has instead treated a spiking neuron as a stochastically firing unit, where spikes are distributed according to an underlying instantaneous firing rate, that in turn has a smooth dependence on network parameters; for example, in multilayer \ac{ReSuMe} a linear Poisson neuron model was used as a substitute for deterministic spiking neurons in each layer, during its derivation \cite{Sporea2013}.

Multilayer learning rules have demonstrated success on several benchmark classification tests, including the linearly nonseparable XOR computation and Iris dataset \cite{Bohte2004,Sporea2013} that cannot otherwise be solved by single-layer networks. However, aside from the work of \cite{Sporea2013}, no attempts have been made in establishing the performance of a multilayer spiking network when learning to perform a large number of input-output spike pattern mappings; it is likely that the presence of more than one layer can enhance the storage capacity of the network, by increasing the number of spiking neurons that can perform computations on network inputs. Progress in this area has been hindered by the complexity that arises from applying learning rules to multilayer spiking networks.

Much of the previous work examining the performance of both single- and multilayer learning rules for spiking networks have considered simplified coding schemes. For example, both SpikeProp and the Chronotron have used the latency of single output spikes to encode for different input patterns, and the Tempotron used a binary spike / no-spike output code to discriminate between two classes of inputs. Ideally, for spiking networks a fully temporal coding scheme would be taken advantage of such that input patterns were encoded by the precise timing of multiple output spikes. We have previously indicated the advantages of using a fully temporal code in \cite{Gardner2014a}, and in particular found that multiple, rather than single output spikes, increased the reliability of classifications.

Most learning rules have been applied to networks containing just a single output neuron. Biologically, however, it is well known that populations of neurons encode for similar patterns of activity, such that the detrimental impact of synaptic noise on neural processing can be eliminated \cite{Faisal2008}. In a series of notable studies \cite{Urbanczik2009,Friedrich2010,Friedrich2011}, groups of spiking neurons receiving shared input patterns were simulated to mimic such a population-based coding scheme: with the key result that the speed of learning increased with the population size. Such studies were devised in the framework of reinforcement learning and typically used a spike / no-spike or latency code to classify input patterns; hence, it would be of interest to investigate populations of spiking neurons utilizing a fully temporal code with multiple output spikes.

Here we derive a supervised learning rule for a multilayer network of spiking neurons which is capable of encoding input spike patterns by the precise timings of multiple output spikes. Our rule extends the single-layer learning rule of Pfister et al. \cite{Pfister2006} to multiple layers by combining the method of stochastic gradient ascent with backpropagation.
We demonstrate the efficacy of the proposed learning rule on a wide variety of learning tasks: both in terms of the accuracy of input pattern classifications and the time taken to converge in learning. We find the learning rule can encode for a large of number of input patterns, comparing favourably with previous multilayer learning rules, and results in increased classification accuracy when classifying inputs by the timings of multiple rather than single output spikes. The learning rule is further applied to multilayer networks containing multiple output neurons, where we measure the dependence of the performance on the specific network setup when mapping between spatio-temporal spike patterns. Finally, we propose a biologically plausible implementation of the multilayer learning rule, and predict the underlying neural mechanisms that might guide the learning of desired target output spike trains.
Our multilayer learning rule differs from those proposed by Brea et al. \cite{Brea2013} and Rezende \& Gerstner \cite{JimenezRezende2014}, which have instead taken gradient descent on the KL-divergence in a supervised and reinforcement setting respectively. The novelty of our paper comes from the application of backpropagation, and its indicated high performance when encoding for a large number of input spike patterns as multiple and precisely timed output spikes.

\section*{Results}
We introduce our learning rule for a feedforward network of spiking neurons containing a hidden layer. The performance of the proposed learning rule is then examined on a variety of benchmark tests: first for multilayer networks containing a single output neuron as the readout, and secondly for multilayer networks containing multiple output neurons. With respect to single-output networks, learning tasks include: measuring the resilience of the network to noise, solving the XOR computation, a comparison over specific network setups, the storage capacity of the network and its ability to generalize on a synthetic dataset. For multiple-output networks, the performance of the learning rule is tested on mapping between multiple input-output spatio-temporal spike patterns, and the ratio of hidden to output neurons required to attain reliable input classifications. Finally, we present an alternative and more biologically plausible formulation of backpropagated learning, and compare its performance against that of our derived rule for both single- and multiple-output multilayer networks.

\subsection*{Learning rule}

\paragraph{Neuron model.} We consider a postsynaptic neuron, indexed $o$, that receives its input from other presynaptic neurons $h$. If the postsynaptic neuron generates a list of spikes $z_o = \{t_o^1, t_o^2, ...\}$ in response to the presynaptic spike pattern $y_h \in \mathbf{y}$, then its membrane potential at time $t$ is defined by the \acf{SRM} \cite{Gerstner2002}:
\begin{equation}
u_o(t) := \sum_h w_{oh} ( \mathcal{Y}_h \ast \epsilon )(t) + ( \mathcal{Z}_o \ast \kappa )(t) \;, 
\end{equation}
where $w_{oh}$ is the synaptic weight between neurons $h$ and $o$, and both $( \mathcal{Y}_h \ast \epsilon )(t)$ and $( \mathcal{Z}_o \ast \kappa )(t)$ denote a convolution between a spike train and a \acf{PSP} kernel $\epsilon$ and reset kernel $\kappa$ respectively (Methods). A spike train is given as a sum of Dirac $\delta$ functions: $\mathcal{Y}_h(t) = \sum_f \delta (t - t_h^f)$, and a convolution is defined by
\begin{equation} \label{eq:convolution}
( \mathcal{Y}_h \ast \epsilon )(t) \equiv \int_0^t \mathcal{Y}_h(t')\, \epsilon(t-t') \mathrm{d}t' \;.
\end{equation}
In our analysis we implement a stochastic neuron model, such that postsynaptic spikes are distributed according to an instantaneous firing rate:
\begin{equation} \label{eq:firing_density}
\rho(t) = g[u(t)] \;,
\end{equation}
where $g[u]$ is a monotonically increasing function of the neuron's membrane potential (see also Eq. \ref{eq:EXP_rate}). 

\paragraph{Supervised learning.} The learning rule is derived for a fully connected feedforward network containing a single hidden layer. Input layer neurons just present spike patterns to the network, while both hidden and output neurons are free to perform computations on their respective inputs. Input layer neurons are indexed as $i \in I$, hidden neurons $h \in H$ and output neurons $o \in O$.

Both hidden and output neurons have their spikes distributed according to Eq. \ref{eq:firing_density}; the advantage of implementing a stochastic neuron model is that it allows for the determination of the likelihood for generating a specific output spike pattern. Hence, if the likelihood of generating a list of target output spikes $z_o^{\mathrm{ref}} = \{\tilde{t}_o^1, \tilde{t}_o^2, ...\}$ in response to $\mathbf{y}$ is $P(z_o^{\mathrm{ref}}|\mathbf{y})$, then the likelihood of generating a spatio-temporal target pattern $z_o^{\mathrm{ref}} \in \mathbf{z}^{\mathrm{ref}}$ is given by the product $P(\mathbf{z}^{\mathrm{ref}}|\mathbf{y}) = \prod_o P( z_o^{\mathrm{ref}}|\mathbf{y})$ with log-likelihood (Methods):
\begin{equation} \label{eq:log_likelihood}
\log P(\mathbf{z}^{\mathrm{ref}}|\mathbf{y}) = \sum_o \int_0^T \log ( \rho_o(t) ) \mathcal{Z}_o^{\mathrm{ref}}(t) - \rho_o(t) \mathrm{d}t \;,
\end{equation}
where $\mathcal{Z}_o^{\mathrm{ref}}(t) = \sum_f \delta (t - \tilde{t}_o^f)$, $T$ the duration over which $\mathbf{y}$ is presented and $\rho_o$ the output firing rate.
We aim to maximize the log-likelihood of generating a target output spike pattern by taking gradient ascent with respect to synaptic weights in the network. For clarity, we just consider a network containing a single hidden layer, although our technique can straightforwardly be extended to include multiple hidden layers. 

\begin{figure}[t!]
\includegraphics{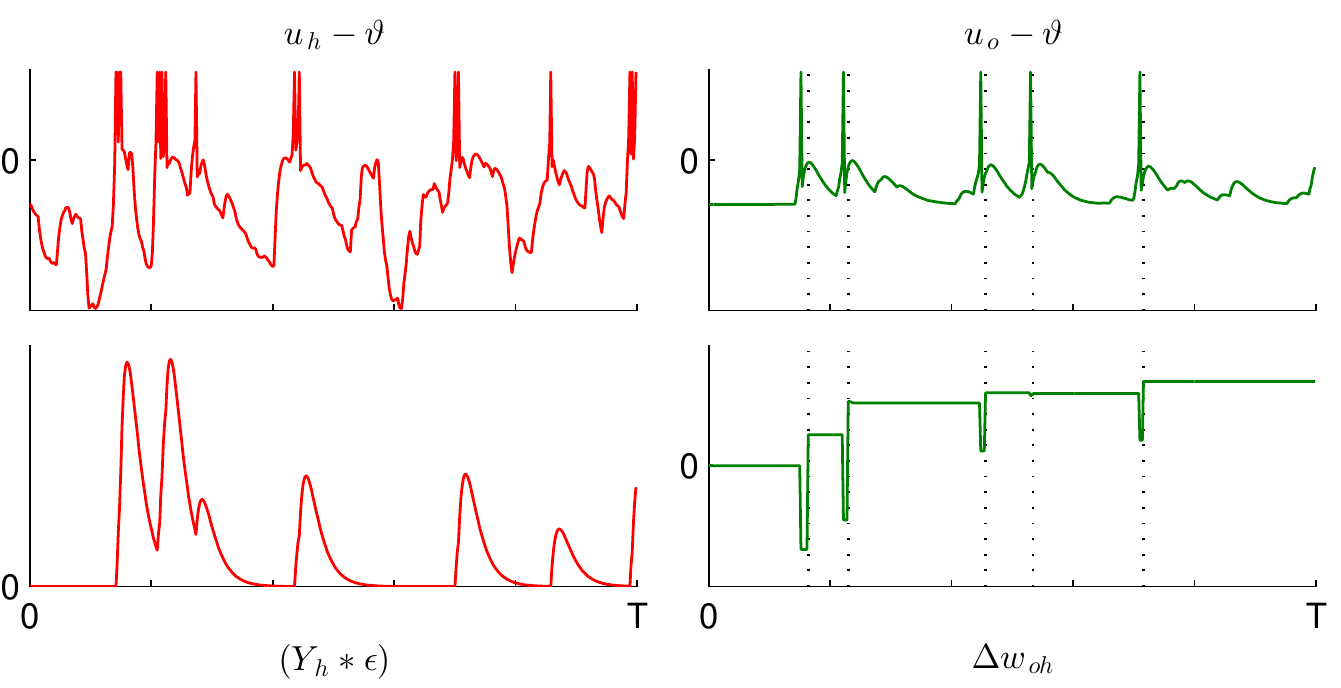}
\caption{
{\bf Example of the output layer weight update rule, in terms of hidden and output neuron activity.}
\textit{Top row:} The left panel shows the fluctuations of a hidden neuron membrane potential relative to a firing threshold $\vartheta$, in response to an input pattern lasting duration $T$, where hidden spike times are indicated by vertical lines. Note that thicker lines indicate phasic bursting, that evoke stronger responses in output layer neurons. The right panel shows the membrane potential of an output neuron, that responds to stimulation from hidden layer neurons. In this example, the output neuron must learn to generate spikes at the times indicated by the dotted lines. 
\textit{Bottom row:} The left panel is the \ac{PSP} evoked at output layer neurons due to hidden neuron spikes, that is defined by Eq. \ref{eq:convolution}. The right panel is the candidate weight change between the hidden and output neurons shown in this example, that depends on both the hidden-evoked \ac{PSP} and the accuracy of the output activity, according to Eq. \ref{eq:output_rule}. Note the depressions in $\Delta w_{oh}$ correspond to the timings of actual output spikes, that are slightly too early with respect their targets, while the increases take place at the timings of target output spikes, subject to the output error signal $\delta_o$ (Eq. \ref{eq:backprop_error}). In this case, the final update $\Delta w_{oh}$ at time $T$ is positive, that demonstrates the causal role of the hidden neuron spikes in eliciting accurate output spike times.
}
\label{fig1}
\end{figure}

From taking gradient ascent on Eq. \ref{eq:log_likelihood} (Methods), the output layer weight update rule is determined as
\begin{equation} \label{eq:output_rule}
\Delta w_{oh} = \eta_o \int_0^T \delta_o(t)\, ( \mathcal{Y}_h \ast \epsilon )(t)\, \mathrm{d}t \;,
\end{equation}
where $\eta_o$ is the output learning rate and $\delta_o$ an output neuron error signal. This error signal measures the dissimilarity between a target response $z_o^{\mathrm{ref}}$ and the actual output activity $\rho_o$, that is given by
\begin{equation} \label{eq:error_signal}
\delta_{o}(t) = \frac{1}{\Delta u_o} \left[ \mathcal{Z}_o^{\mathrm{ref}}(t) - \rho_o(t) \right] \;,
\end{equation}
where $\Delta u_o$ is a parameter that controls the variability of output spike times (Eq. \ref{eq:EXP_rate}). From the above, we find positive values for $\delta_o$ signal the timings of desired output spikes, while negative values signal erroneous output activity. The above learning rule was originally derived by Pfister et al. \cite{Pfister2006} for a single-layer network, that has been found to well approximate the functional form of \ac{STDP} observed experimentally in \cite{Bi1998}. An example of a weight update taking place in the output layer is shown in Fig. \ref{fig1}. 

\begin{figure}[t!]
\includegraphics{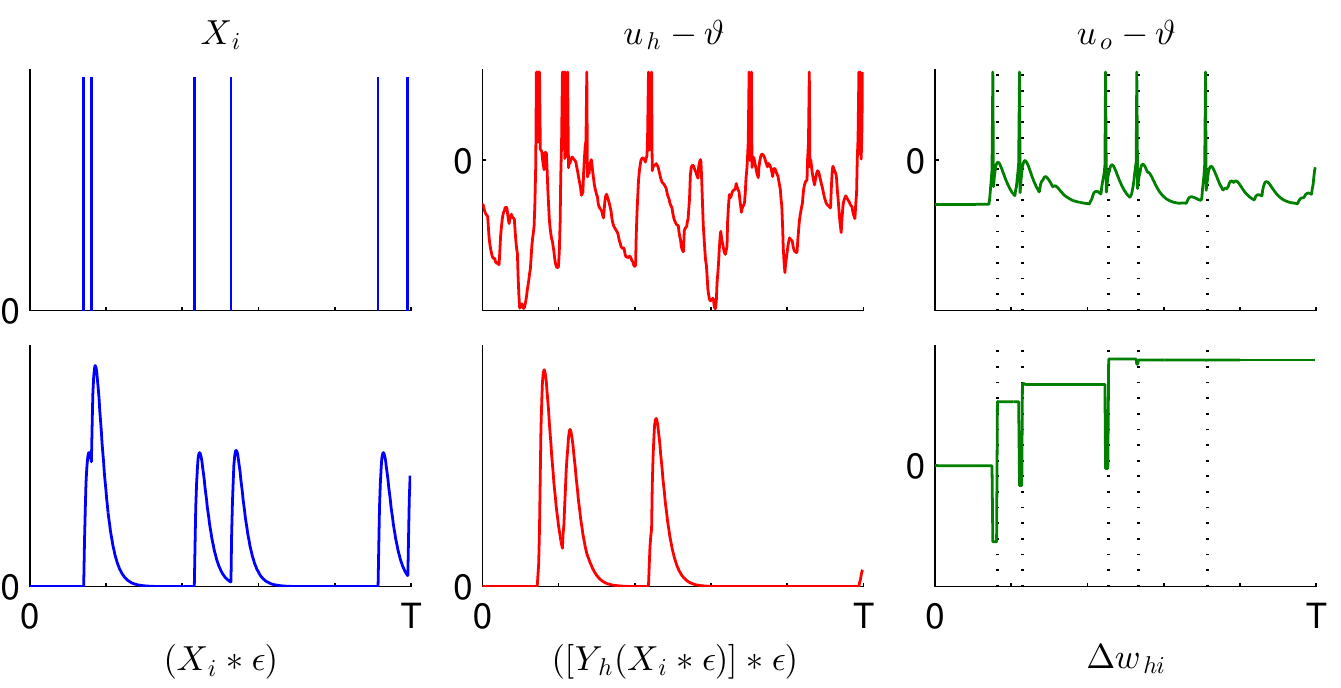}
\caption{
{\bf Example of the hidden layer weight update rule, in terms of input, hidden and output neuron activity.} This figure has the same setup as in Fig. \ref{fig1}.
\textit{Top row:} The left panel shows an input neuron spike train, that is distributed according to a Poisson process. The middle panel shows the membrane potential of a selected hidden neuron, which is partly stimulated by input spikes from the preceding panel. The right panel shows the membrane potential of an output neuron, with target output spike-timings indicated by dotted lines.
\textit{Bottom row:} Shown along this row are a series of synaptic traces, that have a functional dependence on the last panel. The left panel is the \ac{PSP} evoked at hidden layer neurons, due to the input spike train above. The middle panel is a double convolution (Eq. \ref{eq:double_convolution}), that captures the correlation between input and hidden neuron spike-timing. In this case, only input and hidden neuron spikes up to the first half of $T$ are correlated. The right panel shows the progression of the candidate weight change $\Delta w_{hi}$ between the input and hidden neurons selected in this example, that depends on the synaptic trace shown in the middle panel and the accuracy of output spikes, according to Eq. \ref{eq:hidden_rule}. In this case, the final update $\Delta w_{hi}$ effected at time $T$ is positive: demonstrating the causal role of the input spike train in driving an accurate output spike train.
}
\label{fig2}
\end{figure}

By taking gradient ascent on Eq. \ref{eq:log_likelihood} and using the technique of backpropagation (Methods), the hidden layer weight update rule is found as
\begin{equation} \label{eq:hidden_rule}
\Delta w_{hi} = 
\frac{\eta_h}{\Delta u_h} 
\sum_o w_{oh} \int_0^T
\delta_o(t) ([ \mathcal{Y}_h( \mathcal{X}_i \ast \epsilon )] \ast \epsilon)(t) \mathrm{d}t \;,
\end{equation}
where $\eta_h$ is the hidden learning rate, $\Delta u_h$ a parameter that controls hidden neuron spiking variability and $([ \mathcal{Y}_h( \mathcal{X}_i \ast \epsilon )]$ denotes a double convolution (Eq. \ref{eq:double_convolution}). An example of a weight update taking place in the hidden layer is shown in Fig. \ref{fig2}. Given the dependence of weight updates on the availability of hidden neuron spike, it is necessary that a degree of variable activity persists in the hidden layer: an absence of hidden activity would otherwise prevent updates from taking place and result in stagnated learning. To this end, hidden weights are additively modified through synaptic scaling (Methods).

\subsection*{Network setup}

The multilayer learning rule was tested in simulations of networks of stochastic \ac{LIF} neurons, that performed temporally precise input-output spike pattern mappings. In all simulations, input patterns were represented by the firing times of $n_i = 100$ input layer neurons, where an input pattern consisted of a Poisson spike train at each input neuron with a mean firing rate of \SI{6}{Hz} (Methods). Input patterns were presented episodically to the network in no particular order, and weight changes were applied at the end of each episode. Depending on the learning task, a variable number $n_h$ of hidden neurons were implemented in the network to establish the dependence of the performance on the hidden layer size. Here we first present results from simulations of a multilayer network containing a single output neuron as its readout, and then extend our analysis to include a network containing multiple output neurons. For each experiment, a more detailed description of the network setup can be found in the Methods section.

\subsection*{Performance of the learning rule}

The performance of the learning rule is demonstrated by training a multilayer network to perform generic input-output spike pattern mappings. We first focus on the relatively simple task of performing a single input-output mapping, and then extend our analysis to more complex multiple input-output mappings that are subject to noise.

\begin{figure}[t!]
\includegraphics{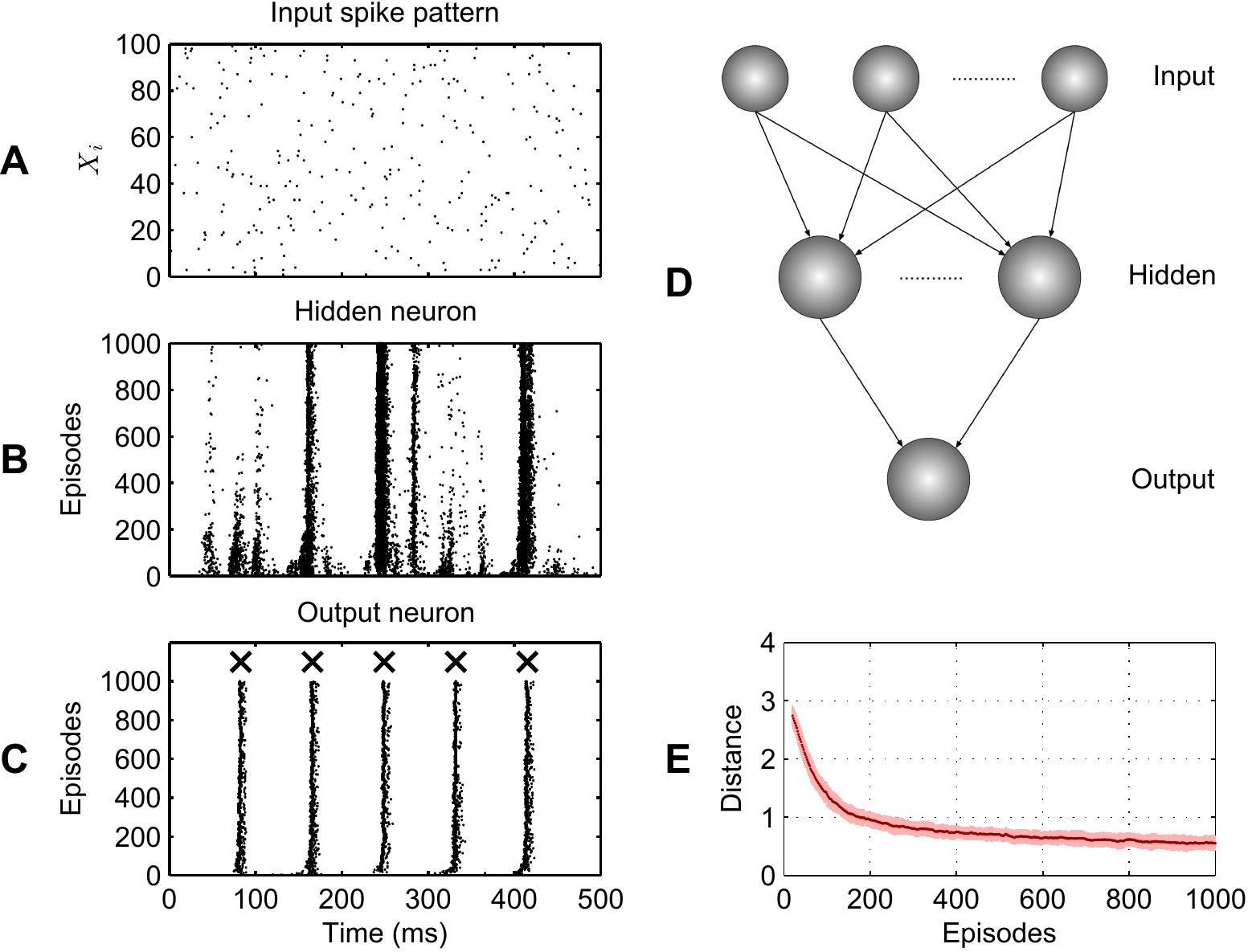}
\caption{
{\bf Learning a target output spike train in response to a single, fixed input pattern.}
The network contained $n_i = 100$ input neurons, $n_h = 10$ hidden neurons and a single output neuron. The input pattern was repeatedly presented to the network over 1000 episodes, where each episode lasted duration $T = \SI{500}{ms}$. The target output spike train contained five spikes at times: \SIlist[list-units = single]{83; 166; 249; 332; 415}{ms}.
(A) A spike raster of the input pattern that was presented to the network on each episode. 
(B) The activity of a hidden neuron with each episode, that contributed strongly to the firing times of the output neuron. 
(C) The activity of the output, where the five target output spike times are indicated by crosses.
(D) An illustration of the multilayer network setup. 
(E) The evolution of the distance between the actual output and target output spike trains of the network, given as a moving average of the \acl{vRD} $\tilde{\mathcal{D}}$ with each episode (Methods) and taken over 100 independent simulation runs. The shaded region shows the standard deviation.
}
\label{fig3}
\end{figure}

\paragraph{Single input-output mapping.} A multilayer network was trained to map between an input pattern and a target output spike train. The network contained 10 hidden neurons and a single output neuron, which was tasked with learning the timings of five target output spikes. An illustration of the network setup is shown in Fig. \ref{fig3}, along with example spike rasters depicting input, hidden and output neuron spiking activity over a typical simulation run. 

In this example, we examined a hidden neuron that contributed strongly to the responses of the output neuron close to the target spike times: \SIlist[list-units = single]{166; 249; 415}{ms} (Fig. \ref{fig3}B). From this hidden neuron spike raster, highly variable spike times were observed over the first 200 episodes, that subsequently fine-tuned themselves to the timings of target output spikes; this initial phase of variable activity demonstrated a form of stochastic exploration by the network, during which time desirable hidden spike patterns were discovered by the network which contributed to accurate output spike times. As learning progressed, hidden neurons generated bursts of spikes around the timings of target output spikes, such that the likelihood of evoking accurate output responses was increased. In this simulation, the majority of hidden layer neurons contributed to driving accurate output spiking responses, hence the load imposed on the network in the form of hidden synaptic modifications was more evenly distributed amongst them.

From the output spike raster (Fig. \ref{fig3}C) it is clear that every target output spike was learnt successfully, and within just 100 episodes. However, because a stochastic rather than a deterministic neuron model was implemented, a small degree of variation in the timings of output spikes about their respective targets was apparent. Despite this, the network still generated output responses to a sufficiently high level of accuracy, that is supported by the \acf{vRD} measure (defined in Eq. \ref{eq:vRD}) with a final average value $\tilde{\mathcal{D}} = 0.55 \pm 0.13$ (Fig. \ref{fig3}E). For an impression of this \ac{vRD} value, a distance of 0.55 corresponds to a typical time shift of \SI{1.17}{ms} between paired actual and target output spikes.

\begin{figure}[t!]
\includegraphics{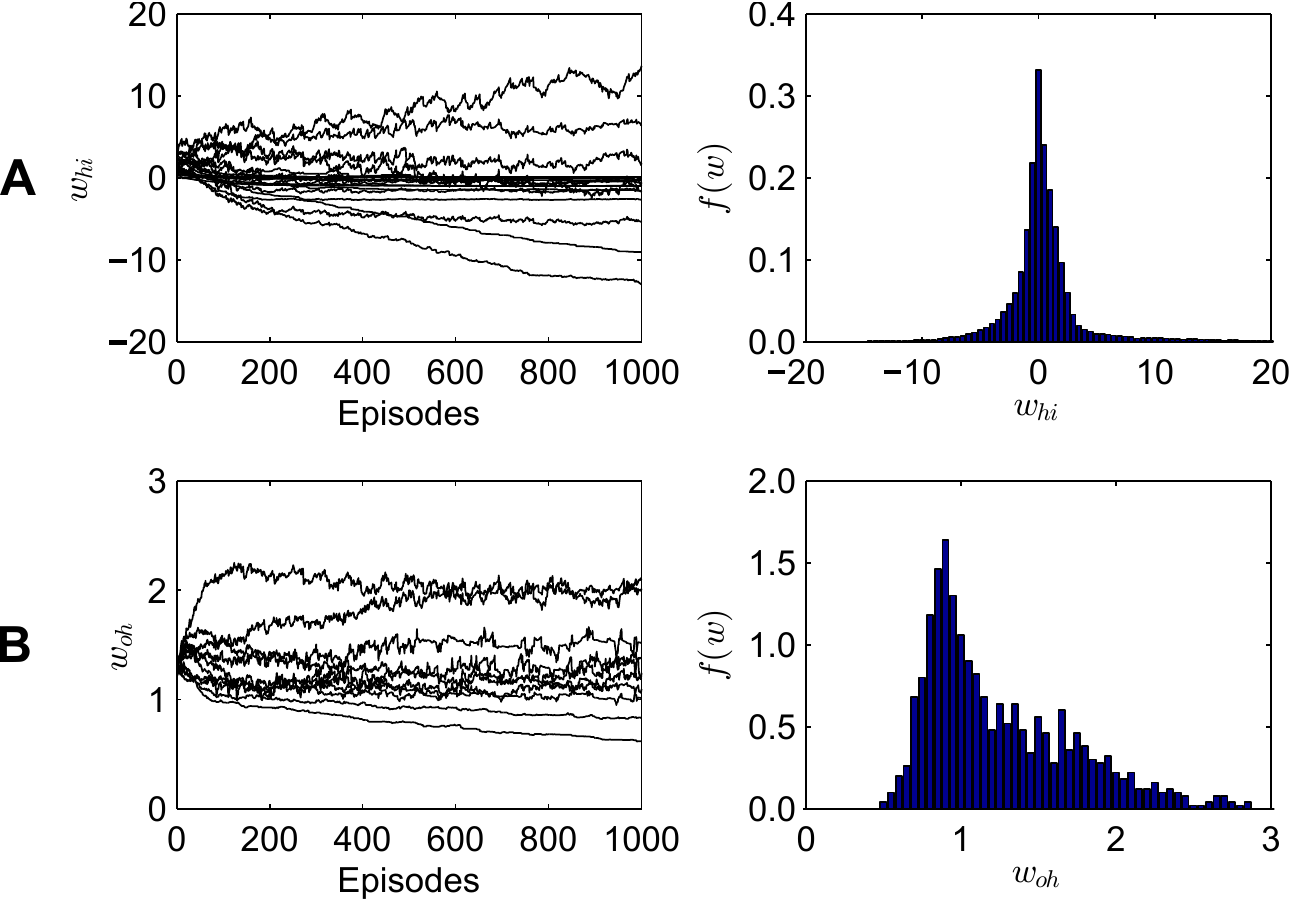}
\caption{
{\bf Evolution and distribution of synaptic weights in a multilayer network, for the simulation shown in Fig. \ref{fig3}.}
\textit{Left column:} An example of the evolution of synaptic weights with the number of learning episodes.
\textit{Right column:} The final distribution of synaptic weights $f(w)$ after 1000 learning episodes.
The panels in row (A) correspond to hidden layer weights $w_{hi}$: the left panel shows the evolution of the first 20 weights on the hidden neuron shown in Fig. \ref{fig3}, and the right panel shows the final distribution over all hidden layer weights.
The panels in row (B) show output neuron weights $w_{oh}$: the left panel shows the evolution of all 10 weights on the output neuron shown in Fig. \ref{fig3}, and the right panel shows the final distribution of output weights.
For both panels showing the final distribution of weights, 100 independent simulation runs were taken.
}
\label{fig4}
\end{figure}

\paragraph{Synaptic weight distributions.}
Shown in Fig. \ref{fig4} is an example of the evolution of both hidden and output synaptic weights with the number of learning episodes and their final distribution, that corresponds to the previous experimental setup. In the left panel (Fig. \ref{fig4}A), the weights on the hidden neuron can be seen to diverge continuously during learning, with almost twice as many positive as negative weights by the final episode. This contrasts with the evolution of the weights on the output neuron (Fig. \ref{fig4}B, left panel), which attained rapid convergence during learning. We note that in our implementation output weights were confined to positive values, while hidden weights had no such restriction (Methods); preliminary simulations indicated that negative output weight values for a single output neuron had little impact on its performance. 

At the end of learning, hidden weights closely followed a Gaussian distribution (Fig. \ref{fig4}A, right panel) and output weights a positively skewed distribution (Fig. \ref{fig4}B, right panel), with coefficients of variation $1.52 \pm 0.01$ and $0.375 \pm 0.009$ in the magnitude of hidden and output weight values respectively. Hence, in terms of the absolute value, hidden weights were more widely dispersed than output weights by a factor of just over four.

\paragraph{Multiple input-output mappings with noise.}
We next tested the performance of the multilayer network when learning to map between 10 input-output spike pattern pairs and the impact of input noise on learning. In this case, each input pattern was identified by a unique target output spike time. The network contained $n_h=10$ hidden neurons and a single output neuron. In this experiment, we introduced two new measures: the time shift $\Delta t$ and the performance $\tilde{\mathcal{P}}_c$. The time shift was taken as a moving average of the absolute difference between matching actual and target output spikes: $\Delta t = |t_o - \tilde{t}_o|$ with each episode, which was computed only for instances when exactly one actual output spike was generated which provided a correct input classification. The measure $\tilde{\mathcal{P}}_c$ was taken as a moving average of the network classification performance (Methods). The time shift $\Delta t$ shared the same averaging window as for $\tilde{\mathcal{P}}_c$, and its motivation came from providing a more physical perspective of the spike train dissimilarity measure $\mathcal{D}$. The performance $\tilde{\mathcal{P}}_c$ measured the accuracy of network classifications based on a temporal code, as described in the Methods section.

\begin{figure}[t!]
\includegraphics{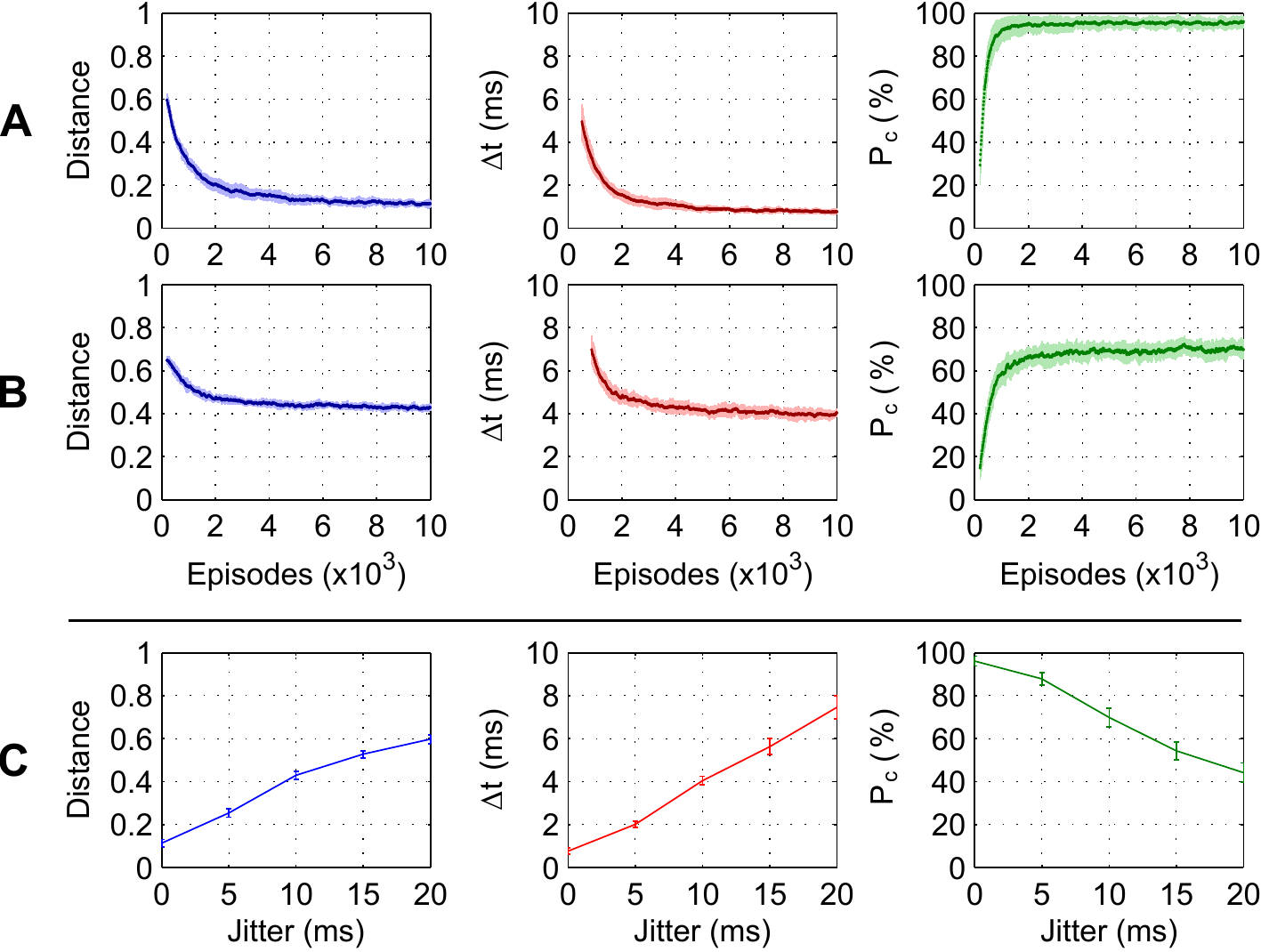}
\caption{
{\bf Learning to map between 10 input-output pattern pairs, with and without input noise.}
The network contained $n_h = 10$ hidden neurons and a single output neuron. Each input pattern was associated with a unique target output spike.
\textit{Left column:} The \ac{vRD} between actual and target output spike trains.
\textit{Middle column:} The time shift between matching actual and target output spikes.
\textit{Right column:} The performance $\tilde{\mathcal{P}_c}$ of the network (Methods), when recognizing input patterns by the timing of an output spike.
(A) Learning in the absence of any input noise, and (B) learning with intermediate input noise. Input noise was simulated by adding jitter to the timings of input spikes on each episode, where jitter with an amplitude of \SI{10}{ms} was used in (B).
(C) Averaged values after $10^4$ learning episodes, as a function of the input jitter amplitude.
In all panels, each value was averaged over 20 independent runs, and error runs show the standard deviation.
}
\label{fig5}
\end{figure}

As shown in Fig. \ref{fig5} learning took place over $10^4$ episodes to ensure convergence, where an input-output pattern pair was randomly selected and presented to the network on each episode. Noise was introduced to the network by jittering the timing of each input spike according to a Gaussian distribution at the start of every episode, with a standard deviation or amplitude that ranged in value from between \SIlist[list-units = single]{0; 20}{ms}.

From the row of panels in Fig. \ref{fig5}A we found that noiseless input patterns resulted in the most accurate output spike times, providing a final distance of $0.11 \pm 0.02$ and a typical time shift of \SI{0.8 \pm 0.1}{ms}. By comparison, introducing \SI{10}{ms} amplitude of input jitter (Fig. \ref{fig5}B) gave a final distance of $0.43 \pm 0.02$ and resulted in output spikes shifted by \SI{4.0 \pm 0.2}{ms}, thereby reducing the temporal precision of output spikes by a factor of five. In terms of the accuracy of input classifications, noiseless inputs resulted in a high performance level of \SI{96 \pm 2}{\%}, which dropped to \SI{70 \pm 4}{\%} with the addition of \SI{10}{ms} amplitude of input jitter. Input noise increased the time taken to converge in learning, taking \num{1.5 \pm 0.3 e3} and \num{2.0 \pm 0.2 e3} episodes for noiseless and noisy (\SI{10}{ms} jitter) inputs respectively.

The panels in Fig. \ref{fig5}C summarise results obtained for \SIlist[list-units = single]{0; 5; 10; 15; 20}{ms} amplitude of input jitter, which show a smooth decrease in the network performance with the degree of input noise. However, even for up to \SI{20}{ms} amplitude of input jitter output spikes still fell within \SI{10}{ms} of their targets and inputs were classified correctly at least \SI{40}{\%} of the time. This remains well above the chance performance level of \SI{10}{\%}, thereby demonstrating the robustness of the multilayer network to strong input noise.

The learning rule has proven capable of training a multilayer network to perform generic input-output spike pattern mappings, and in particular when applied for inputs subject to a high level of noise. We have also indicated the necessity of both active and variable hidden neuronal spiking to ensure convergence of the learning rule, which was supported through synaptic scaling of hidden weights. Next, we examine in more detail the advantages of introducing a hidden layer, and compare the performance of our multilayer learning rule against that for a single-layer network.

\subsection*{Dependence on network structure}

In this section we compare the performance of multi- and single-layer networks as applied to an example classification task, and when performing an increasing number of arbitrary input-output spike pattern mappings. The aim is to support the validity of our multilayer learning rule as an efficient neural classifier.
\paragraph{The XOR computation.}
The learning rule was applied to solving the exclusive-or (XOR) computation, that is a non-trivial classification task. This is considered a standard benchmark for neural network training, given that a hidden layer is necessary for its solution \cite{Gruning2012}.

An XOR computation maps two binary inputs to a single binary output as follows: $\{0, 0\} \rightarrow 0$, $\{0, 1\} \rightarrow 1$, $\{1, 0\} \rightarrow 1$ and $\{1, 1\} \rightarrow 0$. To represent binary values as spike patterns, we used a similar setup to that in \cite{Gruning2012,Seung2003}. For the inputs, each binary value was encoded by a set of 50 Poisson spike trains with a mean firing rate of \SI{6}{Hz}, predetermined at the start of each simulation run; hence, paired binary input values were represented by spike patterns over two groups of 50 neurons. For the output a latency coding scheme was used, where the binary values 0 and 1 corresponded to late/early output neuron spike-timings of \SI{334}{ms} and \SI{167}{ms} respectively. In our simulations we considered multi- and singlelayer networks: both networks contained 100 input neurons and a single output neuron, and the multilayer network contained 10 hidden neurons. For single-layer networks, Eq. \ref{eq:output_rule} was applied to updating input-output weights. Binary inputs were presented to the network episodically in a random order. A correct classification of an input was made when an actual output spike train was closest to its target output as measured by the \ac{vRD}.

\begin{figure}[t!]
\includegraphics{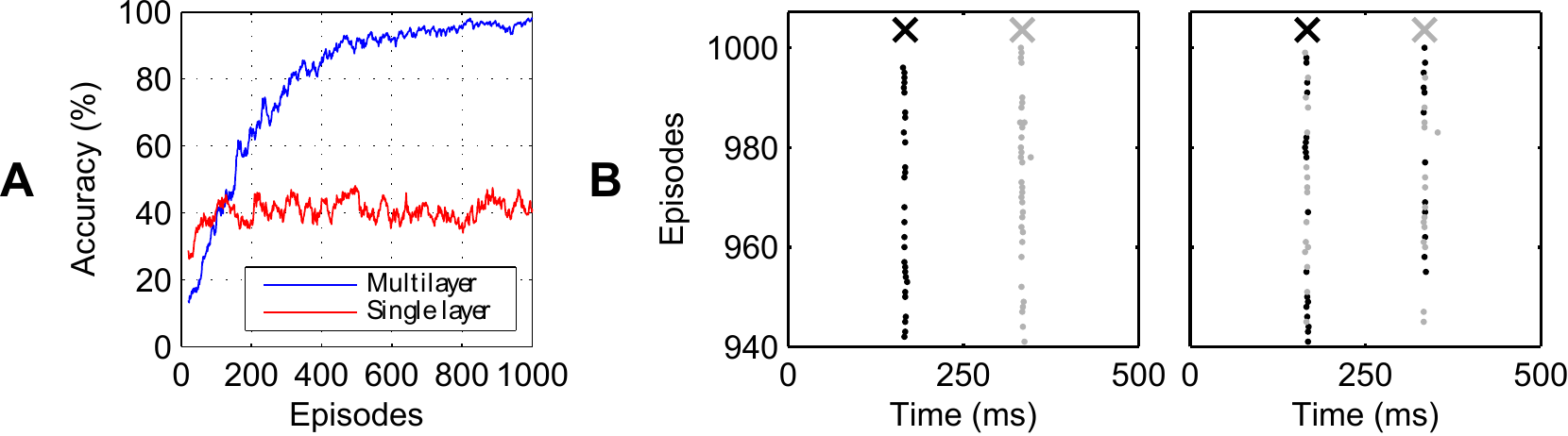}
\caption{
{\bf Learning the XOR function of two binary variables as a temporal code.}
Either a multilayer or single-layer implementation was used. Both networks contained $n_i = 100$ input neurons and a single output neuron. The multilayer network contained $n_h = 10$ hidden neurons. The two binary variables were encoded as predetermined spike patterns over two populations of input neurons, each of size 50. The latency of an output spike coded for the binary value 1 (early spiking) or 0 (late spiking). 
(A) Evolution of the classification accuracy for multi- and single layer networks, averaged over 20 independent runs. 
(B) Output spike rasters for multilayer (left panel) and single layer (right panel) networks, taken over the final 60 episodes on an example run. Black dots correspond to responses from inputs \{0, 1\} and \{1, 0\}, and grey dots correspond to responses from inputs \{0, 0\} and \{1, 1\}. Target spike times are indicated as crosses for each class of input. Results were averaged over 20 independent runs.
}
\label{fig6}
\end{figure}

From Fig. \ref{fig6}A it can be seen that the multilayer network was successful at learning the XOR computation within 1000 episodes, with a final accuracy approaching \SI{100}{\%}. The single-layer network, however, maintained an accuracy around \SI{40}{\%} that is consistent with chance level. It is further apparent from Fig. \ref{fig6}B that the multilayer network was capable of separating the two classes, such that output spike responses for each input class matched their respective targets. In contrast, the single layer network generated erroneous output spikes in response to both input classes, which is indicative of its failure to discriminate between the two classes. Hence, these results support the necessity of including a hidden layer in a spiking network when solving the linearly non-separable XOR computation.

\paragraph{Multiple input-output mappings.}

The performance of various network setups when learning to map between an increasing number of input-output spike patterns was tested. Specifically, the performance of three different network setups were examined: a `free' multilayer network, a `fixed' multilayer network and a single-layer network. Both free and fixed multilayer networks contained 10 hidden neurons and a single output neuron, but differed from each other by their restriction on hidden weight updates: a free multilayer network was allowed changes in the hidden weights during learning by Eq. \ref{eq:hidden_rule}, while hidden weights were not allowed to change in the fixed multilayer network other than through synaptic scaling. The single-layer network lacked a hidden layer and contained a single output neuron. For a more direct comparison, both multi- and single-layer networks contained 100 input neurons.

\begin{figure}[t!]
\includegraphics{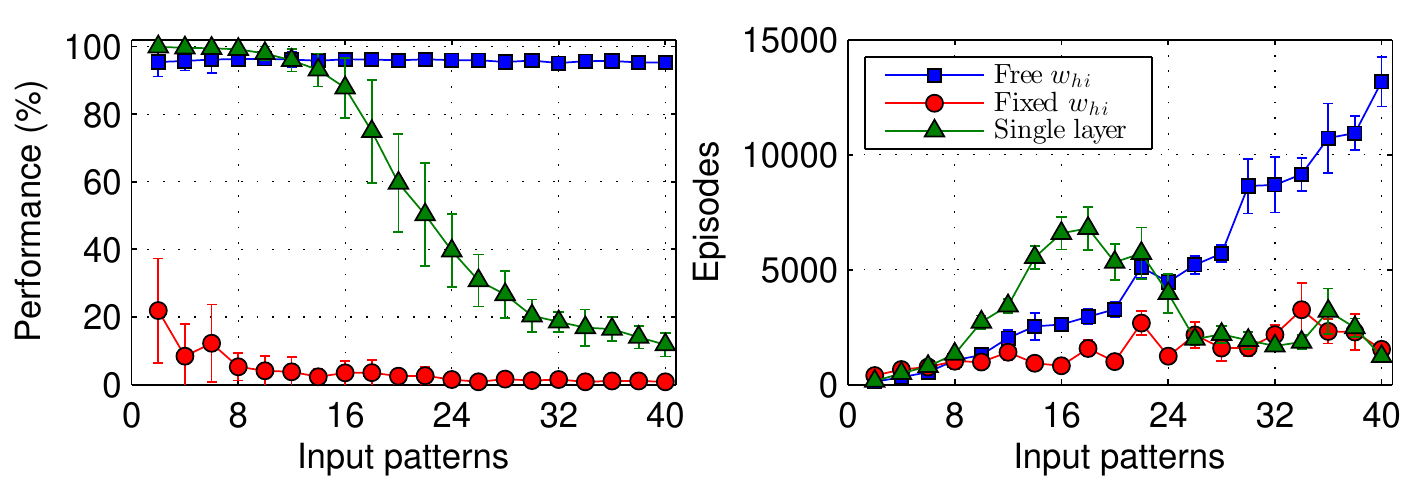}
\caption{
{\bf The dependence of the performance on the number of input patterns and network setup.} Each input pattern was associated with a unique target output spike.
\textit{Left:} The performance as a function of the number of input patterns.
\textit{Right:} The number of episodes to convergence in learning (Methods).
The blue curve shows the performance of a multilayer network where hidden weights $w_{hi}$ are free to be updated according to Eq. \ref{eq:hidden_rule}, and the red curve is a multilayer network with fixed hidden weights. The green curve corresponds to a single-layer network with no hidden layer. Each network contained a single output neuron.
Left panel error bars show the standard deviation, and right panel error bars show the \acf{SEM}; the convergence measure was subject to high variance in most cases, therefore just the average number of episodes taken to converge in learning was considered, and not its distribution. Results were averaged over 20 independent runs.
}
\label{fig7}
\end{figure}

Shown in Fig. \ref{fig7} is the dependence of the network performance on the number of input patterns $p$, up to a maximum of 40, where each input pattern was associated with a unique target output spike. From the left panel, it is clear that both the free multilayer and single-layer networks outperformed the fixed multilayer network over the entire range of input patterns considered; for example, after learning 40 inputs the performance values were \SI{95.2 \pm 0.3}{\%}, \SI{11.8 \pm 0.8}{\%}, and \SI{0.8 \pm 0.2}{\%} for free, single and fixed respectively. The performance of the fixed multilayer network remained consistently low over the entire range of inputs considered, with a maximum value of \SI{22 \pm 3}{\%} for just two inputs. Hence, it is apparent that allowing hidden weight updates to take place is necessary in training a multilayer network to perform input-output pattern mappings. From comparing the free multilayer and single-layer networks, it can be seen that the performance of the single-layer network was greatest for less than 12 inputs; however, for a greater number of inputs the performance of the free multilayer network dominated over the single-layer network. Over the entire range of inputs considered, the performance of the free multilayer network remained around \SI{96}{\%}, and showed no indication of decreasing.

Shown in the right panel are the number of episodes taken for each network to converge in learning (Methods), as a function of the number of inputs. It can be seen that the convergence time for a multilayer network increased with the number of inputs, and was an order of magnitude larger for free in comparison with fixed when learning 40 inputs. The difference in convergence time between the free and fixed multilayer networks was attributed to the increased performance of the free multilayer network: a larger number of episodes was necessary to reach an increased performance value. Finally, the convergence time for a single-layer network decreased when learning more than 18 inputs, which coincided with a rapid drop in its performance level.

To summarise, these results are supportive of multilayer over single-layer learning, and importantly when linearly non-separable classifications are performed for which the presence of a hidden layer is essential. In order for single-layer networks to remain competitive with multilayer networks when mapping between a large number of patterns it would be necessary to scale up the number of input layer neurons, although clearly this would be disadvantageous when more sparse input representations are desired.

\subsection*{Capacity of the multilayer network}

An important consideration when training any neural network is the maximum amount of information it can memorize. Therefore, we measured the dependence of the performance on the number of input patterns that were presented to a multilayer network, that extends the previous experiment in Fig. \ref{fig7}. Given our implementation of a multilayer network, we also explored the dependence of the performance on the hidden layer size. Finally, the dependence of the performance on the number of target output spikes used to identify inputs was tested. The aim was to establish the relationship between the hidden layer size and the number of target output spikes that could be supported, and how this impacted on the network capacity.

\begin{figure}[t!]
\includegraphics{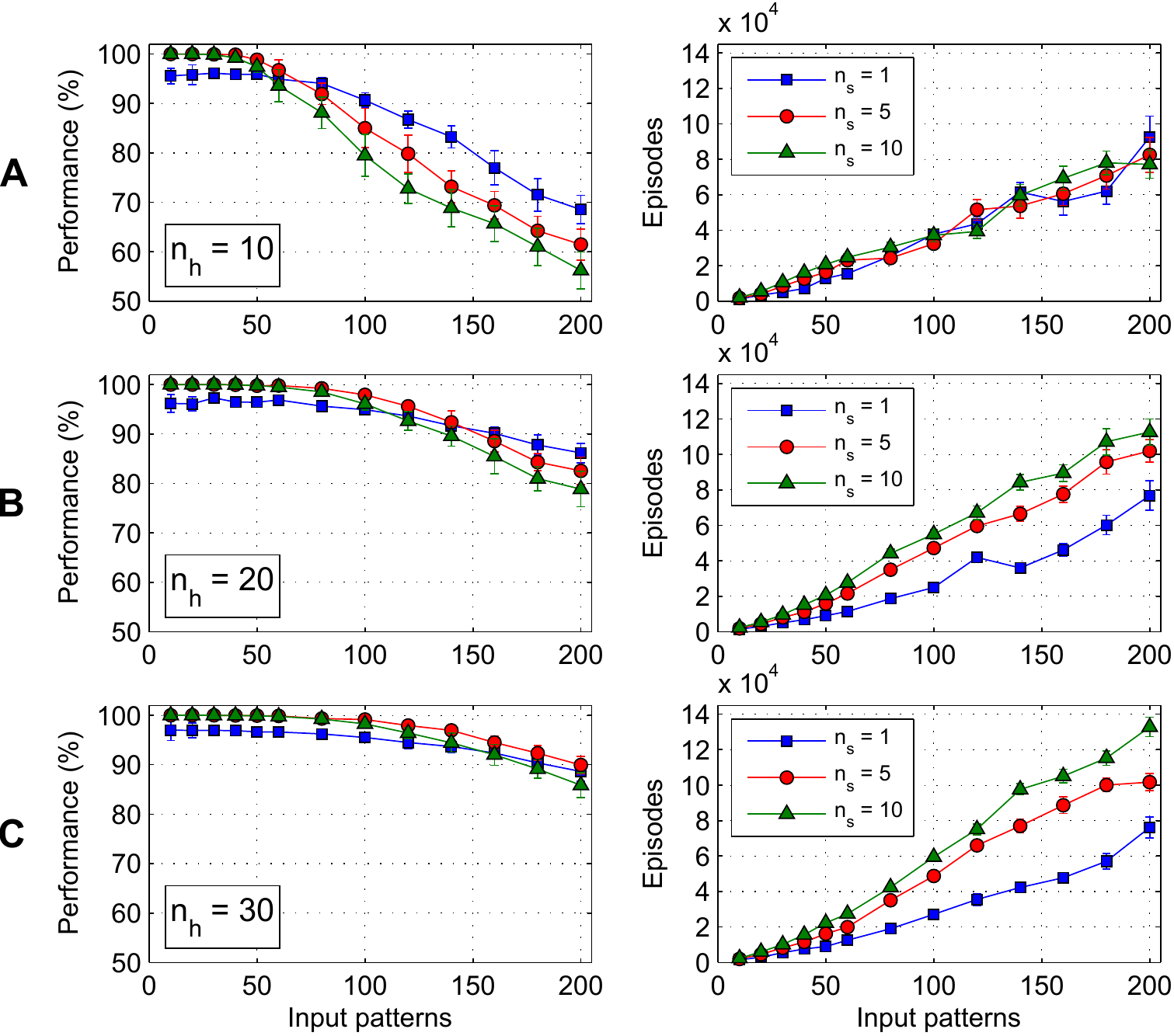}
\caption{
{\bf The dependence of the network performance on the number of input patterns, the number of hidden neurons $n_h$, and the number of target output spikes $n_s$.} In all cases, the network contained one output neuron. In this experiment, input patterns $p$ were equally assigned between $c = 10$ classes.
\textit{Left:} The performance as a function of the number of input patterns, for $n_h = 10$ (A), $n_h = 20$ (B) and $n_h = 30$ (C) hidden neurons. In each panel, different colour curves correspond to the number of target output spikes belonging to each class.
\textit{Right:} The number of episodes to convergence in learning.
Results were averaged over 20 independent runs.
}
\label{fig8}
\end{figure}

In this experiment, the network was tasked with classifying an increasing number of input patterns $p$ into $c = 10$ different classes. An equal number of input patterns were assigned to each class, and all inputs belonging to the same class were identified by a unique target output spike train containing between 1 and 10 spikes. In terms of the network setup, the network contained either 10, 20 or 30 hidden neurons and a single output neuron as the readout.

Fig. \ref{fig8} shows the multilayer performance as a function of the number of input patterns and the number of target output spikes $n_s$ identifying each class of input. From comparing results between the different hidden layer sizes, a larger number of hidden neurons was found to support more target output spikes at a given level of performance. For example, 10 hidden neurons resulted in decreased performance when trained on more than a single output spike, for more than 60 input patterns (Fig. \ref{fig8}A), while 30 hidden neurons resulted in increased performance when trained on at least five output spikes, over the entire range of input patterns considered (Fig. \ref{fig8}C). Furthermore, from a closer inspection of Fig. \ref{fig8}, it can be seen that over a small region of input patterns $p < 50$ the network performance approached \SI{100}{\%} when trained on multiple rather than single output spikes, which was more pronounced for a larger number of hidden neurons. To give an indicator of the network's capacity, the maximum number of input patterns learnt at a performance greater than \SI{90}{\%} was around 100, 150 and 200 for 10, 20 and 30 hidden neurons respectively.

In terms of the time taken by the network to perform input classifications, the number of episodes increased with both the number of hidden neurons and number of output spikes: taking up to \SI{70}{\%} longer for 30 over 10 hidden neurons when trained on 200 input patterns and 10 target output spikes. A decreased number of episodes taken to converge in learning was generally indicative of the networks inability to learn all input patterns.

\begin{figure}[t!]
\includegraphics{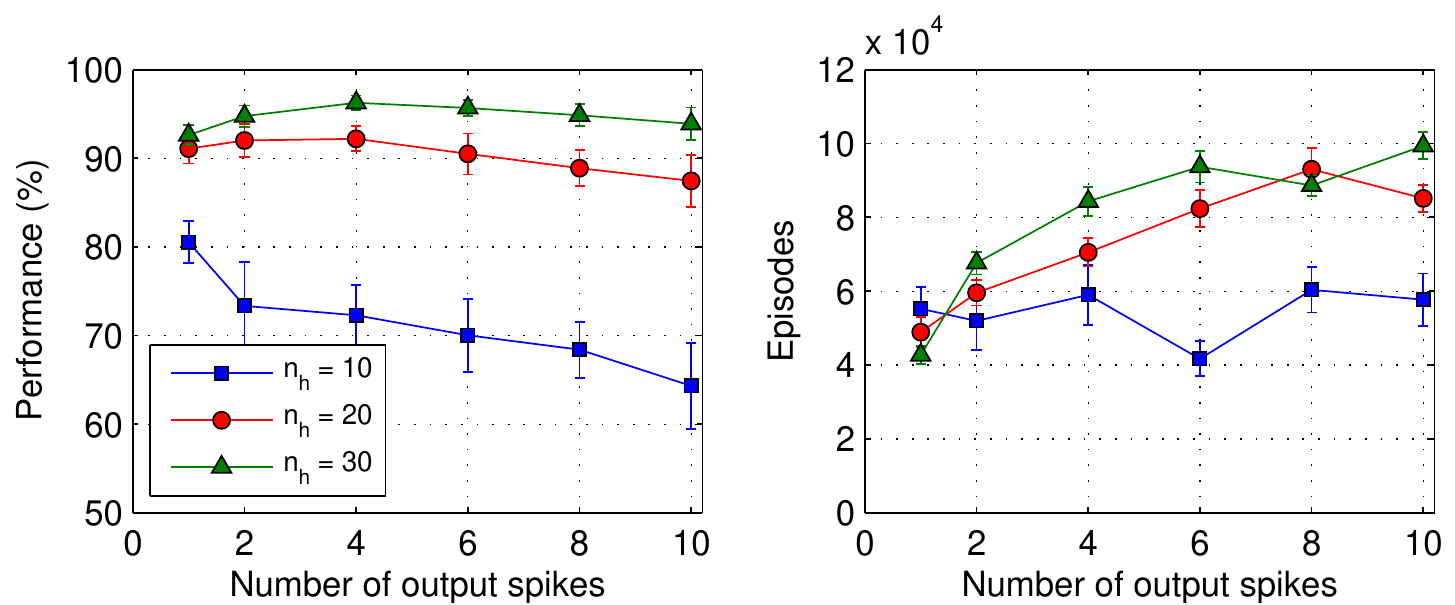}
\caption{
{\bf The dependence of the performance on the number of target output spikes, and the number of hidden layer neurons.} The number of input patterns was $p = 150$ that were equally assigned between $c = 10$ classes, that corresponds to Fig. \ref{fig8}.
\textit{Left:} The performance as a function of the number of target output spikes, for $n_h = 10$, $n_h = 20$ and $n_h = 30$ hidden neurons.
\textit{Right:} The number of episodes to convergence in learning.
Results were averaged over 20 independent runs.
}
\label{fig9}
\end{figure}

In Fig. \ref{fig9} we show in more detail the dependence of the multilayer performance on the number of target output spikes, for different hidden layer sizes. This figure corresponds to the same setup as used in Fig. \ref{fig8} when classifying 150 input patterns into 10 classes. As found previously, a larger number of hidden neurons supported more target output spikes; for example, in the case of 30 hidden neurons, the performance reached a maximum value when trained on around four target output spikes. For just 10 hidden neurons, however, the performance was at its maximum value when trained on just a single target output spike, and decreased as the number of spikes increased. Also evident is an increase in the time taken to learn all inputs for more than 10 hidden neurons and an increasing number of spikes.

From these experiments it is evident that an increase in the hidden layer size provides more capacity to the network, and is supportive of multiple-spike target output trains for more reliable input classifications. These results can be attributed to the internal representations of input patterns afforded by hidden layer neurons, such that class discriminations can be performed at an early stage before being processed by the readout. Qualitatively, it was observed from spike rasters that individual hidden neurons selectively responded to certain input patterns, and only contributed to generating a fraction of the total number of target output spikes. From this, it is apparent that neurons in the hidden layer are capable of distributing the synaptic load imposed upon them, as previously found for the experiment in Fig. \ref{fig3} when performing single input-output mappings.

\subsection*{Generalization ability}

The ability of the network to generalize from stored patterns to similar, new input patterns was tested. Here we built on the earlier experiment (c.f. Fig. \ref{fig5}) which examined the impact of noise on mapping between multiple input-output pattern pairs, by instead considering a more realistic data set that contained several classes of input patterns. The network was tasked with identifying similar inputs belonging to the same class by the timings of output spikes.

We devised a synthetic data set that was inspired by Mohemmed et al. \cite{Mohemmed2012}. Specifically, the accuracy (or classification performance) of the network was tested on a generated dataset that consisted of both training and testing patterns, where the aim of the network was to learn to classify patterns between 10 classes. In generating the training patterns, a single reference spike pattern was randomly created for each class. Each of the 10 reference patterns were then duplicated 15 times, where every duplicate was subsequently jittered according to a Gaussian distribution with a noise amplitude between 2 and \SI{20}{ms}. Hence, a total of 150 training patterns were generated. In the same way, 25 testing patterns were generated for each class, giving a total of 250 testing patterns. Each class of training and testing patterns were associated with a unique target output spike train, that contained between 1 and 5 spikes. On this task, only training patterns were used to train the network, and testing patterns were used to test the ability of the network to generalise. The network contained 20 hidden neurons and a single output neuron.

\begin{figure}[t!]
\includegraphics{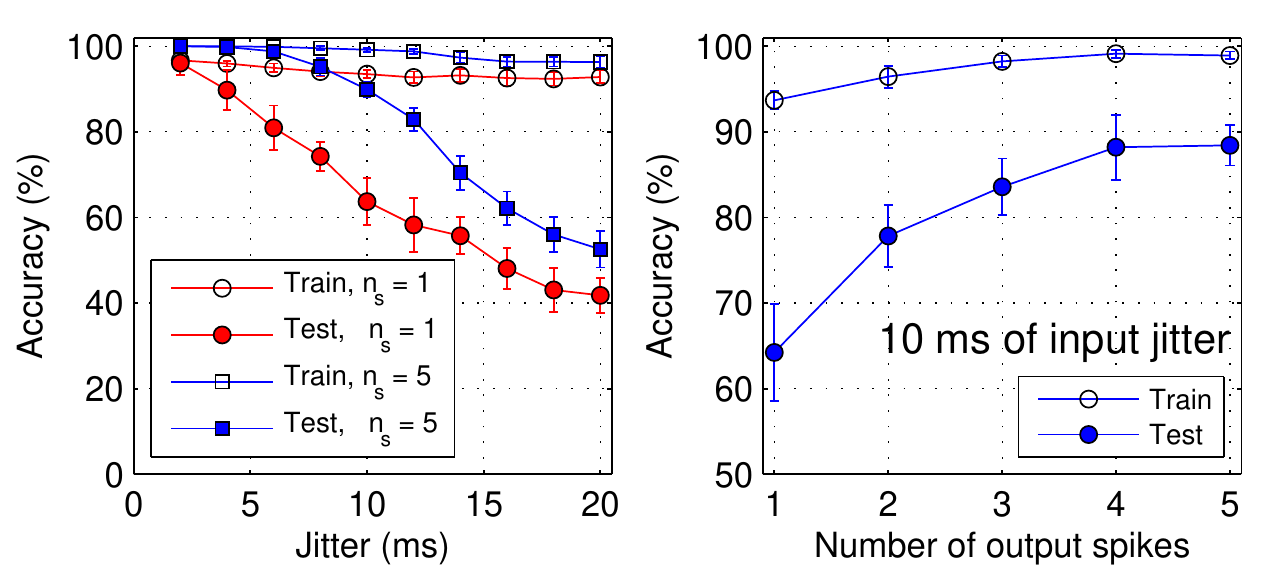}
\caption{
{\bf The classification accuracy on a synthetic dataset, at varying noise levels.}
The network contained $n_h = 20$ hidden neurons and a single output neuron.
The number of training patterns was 150, and the number of testing patterns 250. Both training and testing patterns were equally assigned between 10 classes.
\textit{Left:} The testing and training accuracy as a function of the amplitude of input jitter in the generated dataset, for 5 and 10 target output spikes per class.
\textit{Right:} The testing and training accuracy as a function of the number of target output spikes, at \SI{10}{ms} amplitude input jitter.
To ensure convergence in learning, the number of training episodes was 75000.
Results were averaged over 20 independent runs.
}
\label{fig10}
\end{figure}

Shown in Fig. \ref{fig10} is the network accuracy as a function of the noise amplitude, used to generate input patterns at initialization, and the number of target output spikes. As can be expected, a high degree of noise presented a greater challenge to the network, given that the network had to learn to generalize well in order to accurately classify unseen patterns during the testing phase. Despite this, the network still managed to classify testing patterns at least \SI{40}{\%} of the time at \SI{20}{ms} noise. Furthermore, it is clear that multiple target output spikes led to more accurate classifications in comparison with a single target output spike, giving an increase of almost \SI{25}{\%} at \SI{10}{ms} noise. From the right panel, a smooth increase in the accuracy with the number of target output spikes at \SI{10}{ms} noise can be seen, along with a reduction in the standard deviation; the accuracy of one target output spike was \SI{64 \pm 6}{\%}, compared with \SI{88 \pm 2}{\%} for five output spikes. However, the difference in the accuracy between single and multiple target output spikes became minimal as the noise amplitude approached \SI{20}{ms}.

The network is able to generalize well to similar input patterns, and especially when classifications are performed using multiple target output spikes. Two key reasons explain the increase in accuracy with the number of target spikes. The first relates to the redundancy inherent in multi-spike based classifications: even if an actual output spike train cannot match its target in terms of the number of spikes, an accurate classification can still be performed if those spikes which remain are close enough to their respective targets. The second reason comes from the larger separation between classes as the number of target output spikes increases: class discriminations  made by the network are less prone to error from fluctuating output responses.

\subsection*{Learning spatio-temporal output patterns}

Our learning rules support weight updates in a multilayer network containing more than one output neuron, therefore, we tested the performance of the network when learning to map between spatio-temporal input and output patterns. Here a spatio-temporal output pattern consisted of a unique target spike train at each output neuron, and were combined to identify specific input classes (Methods).

\paragraph{Single input-output mapping.}
First, we considered a mapping between a single input-output pattern pair, where the network was tasked with learning a target spatio-temporal output pattern in response to a single, fixed input pattern. In this experiment, the network contained 20 hidden neurons and three output neurons, where each output was assigned a single, unique target spike time. For multiple output neurons, output weights were allowed to change sign during learning (Methods).

\begin{figure}[t!]
\includegraphics{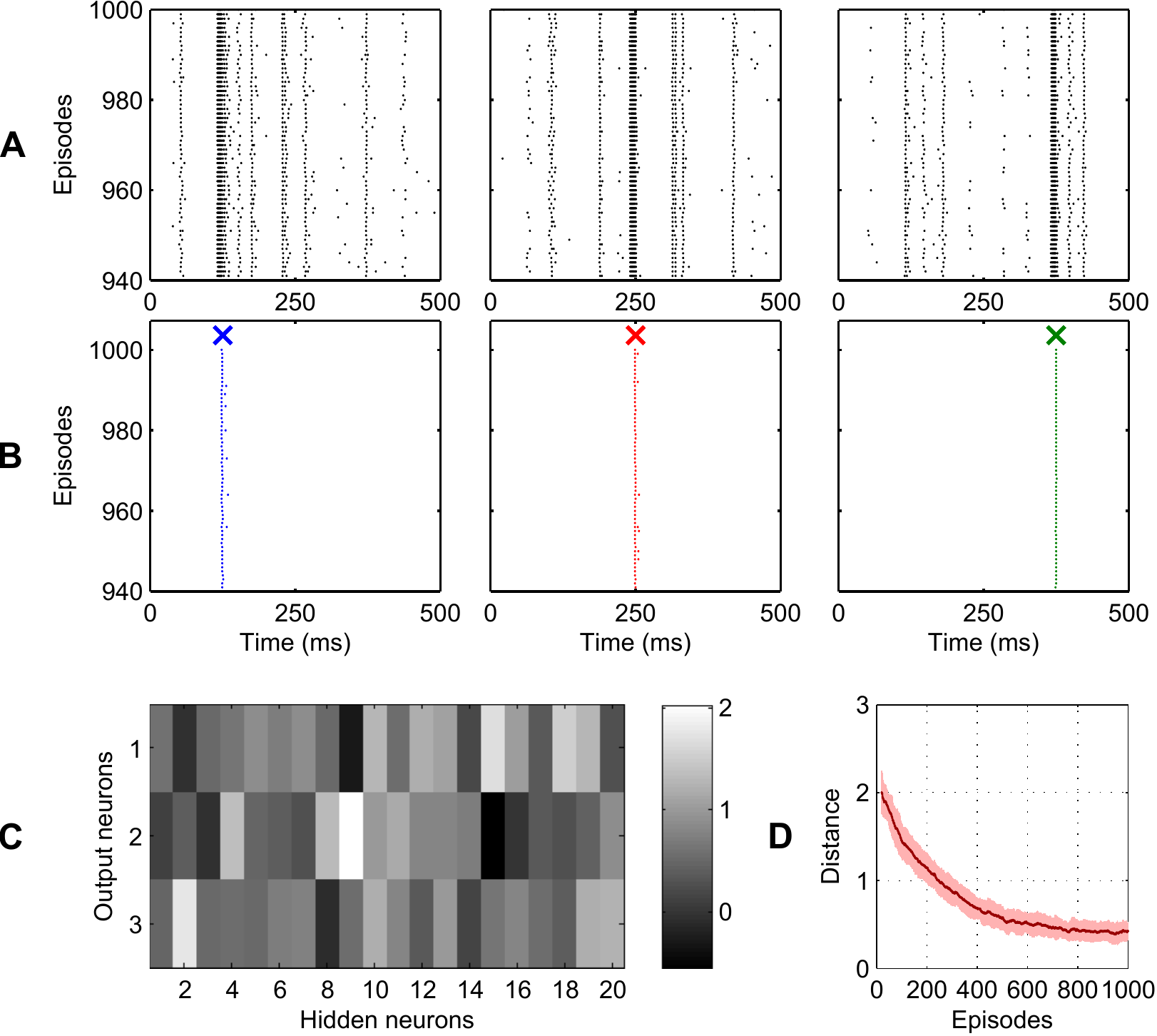}
\caption{
{\bf Learning a mapping between a single input-output pattern pair, in a network containing multiple output neurons.}
The network contained $n_h = 20$ hidden neurons and $n_o = 3$ output neurons. Each output neuron was assigned a unique target spike for the input, at times: \SIlist[list-units = single]{125; 250; 375}{ms} for the first, second and third output neurons respectively. Learning took place over 1000 episodes.
(A) Example hidden neuron spike rasters, and (B) output neuron spike rasters, shown over the final 60 learning episodes.
Each panel in row (A) shows a hidden neuron that contributes strongly to the output neuron response shown in the panel below. In (B), each panel indicates the target spike time of an output with a cross. The left, middle and right panels show the activity of the first, second and third output neurons respectively.
(C) Heatmap of output layer weights $w_{oh}$ after 1000 learning episodes. The intensity corresponds to the strength of synaptic weights. For reference, the left, middle and right panels in (A) show the activity of hidden neuron numbers 15, 9 and 2 respectively.
(D) The evolution of the \ac{vRD}, averaged over 40 independent runs.
}
\label{fig11}
\end{figure}

Fig. \ref{fig11} shows an example of a single simulation run, that depicts hidden (Fig. \ref{fig11}A) and output (Fig. \ref{fig11}B) neuron spike rasters towards the end of learning. Out of the 20 hidden neurons implemented in the network, three were selected for demonstrative purposes that contributed intensely to the target output timings. From Fig. \ref{fig11}A it can be seen that the selected hidden neurons generated stereotypical spike patterns, and particularly around the timings of target output spikes where phasic bursting was observed. In response to hidden layer activity, each output neuron demonstrated a successful learning of their respective target timing (Fig. \ref{fig11}B) and to a good degree of temporal accuracy, that is indicated by a final \ac{vRD} of \num{0.4 \pm 0.1} (Fig. \ref{fig11}D) with a corresponding time shift of \SI{1.9 \pm 0.7}{ms} at each output. Furthermore, the network learnt to distribute the synaptic load between hidden layer neurons, such that hidden spiking activity became more diverse. This is supported by a heatmap of the output weight matrix shown in Fig. \ref{fig11}C, corresponding to the same simulation in panels A and B, which demonstrates a high variance in the synaptic strength between hidden and output neurons.

\paragraph{Dependence on the hidden layer size.}
We next explored the performance of the network when input patterns were classified by  spatio-temporal output patterns. In this experiment, a total of 50 input patterns were equally assigned between 10 classes, such that all five patterns belonging to the same class were identified by a unique, target spatio-temporal output pattern. Target output patterns were randomly predetermined for each class of input. To increase the separation between classes, target output spike trains assigned to each output neuron differed from each other by a \ac{vRD} of at least $n_s / 2$ for $1 \leq n_s \leq 10$ output spikes, similarly as for a network containing a single output neuron. A correct input classification was made when the \ac{vRD} between an actual and desired target output pattern assumed a minimum value (Methods). In measuring the relationship between the performance and network setup, an increasing fractional number $n_h / n_o$ of hidden to output neurons was implemented, for a fixed number of output neurons: $n_o = 10, 20$ or 30. 

\begin{figure}[t!]
\includegraphics{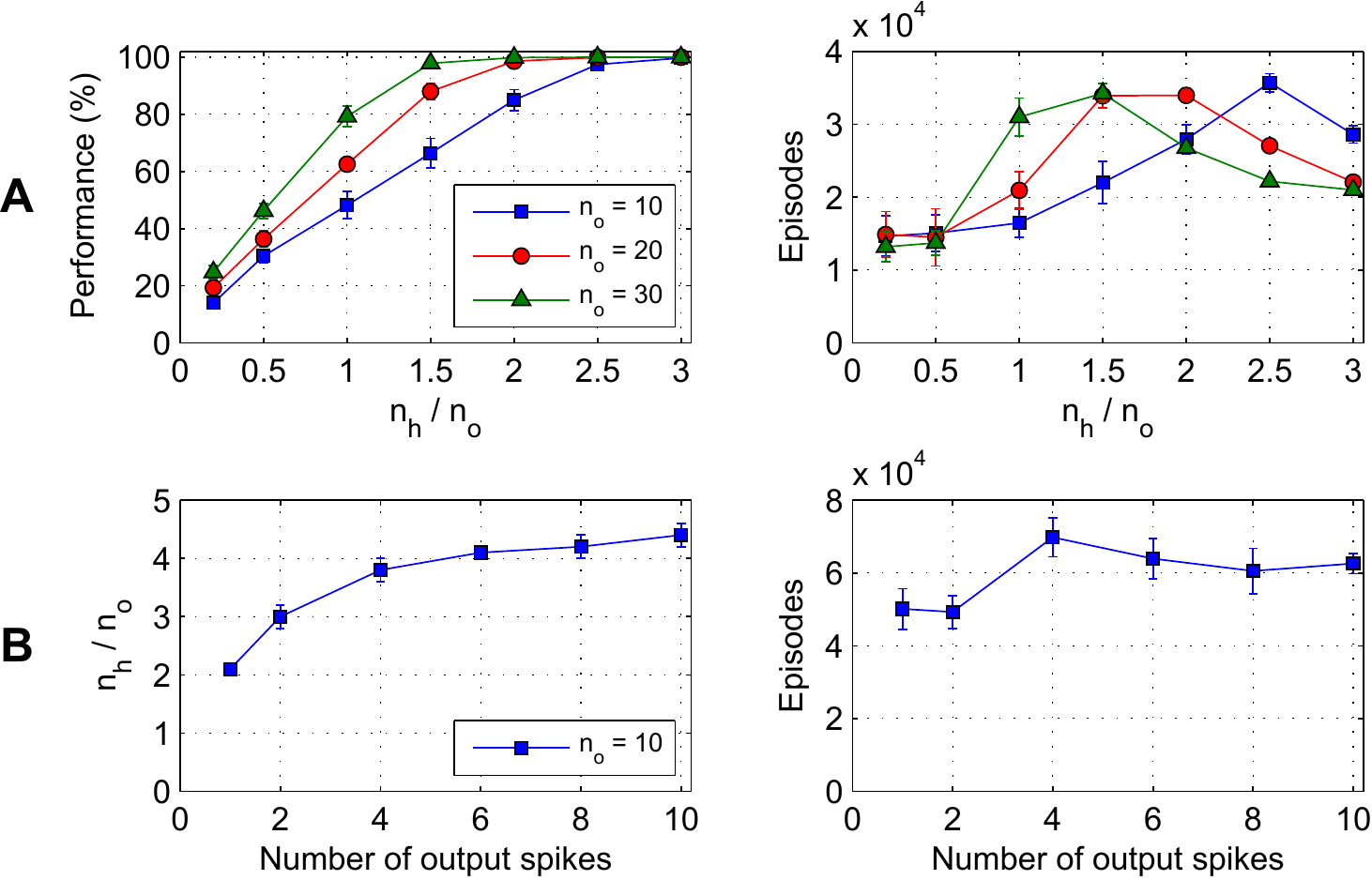}
\caption{
{\bf The dependence of the network performance on the ratio of hidden to output neurons, and the number of target output spikes.}
The network contained an increasing number $n_h$ of hidden neurons, and $n_o = 10, 20$ and $30$ output neurons. $p = 50$ input patterns were equally assigned between $c = 10$ classes, where all patterns belonging to the same class were identified by a unique target output spike pattern.
(A) 
\textit{Left:} The performance as a function of the ratio of hidden to output neurons.
\textit{Right:} The number of episodes to convergence in learning.
(B) 
\textit{Left:} The minimum ratio of hidden to output neurons required to achieve \SI{90}{\%} performance, as a function of the number of target spikes at each output neuron.
\textit{Right:} The number of episodes taken to reach \SI{90}{\%} performance.
Results were averaged over 10 independent runs.
}
\label{fig12}
\end{figure}

From Fig. \ref{fig12}A it is clear that an increase in the fractional number of hidden-output neurons increased the performance of the network, with the performance approaching \SI{100}{\%} for between $2 < n_h / n_o < 3$. Furthermore, there was a dependence of the performance on the number of output neurons; for example, at a fixed fractional number $n_h / n_o = 1$ the performance values were close to \SIlist[list-units = single]{48; 63; 79}{\%} for 10, 20 and 30 output neurons respectively. Hence, it was apparent that a larger number of output neurons increased the separation between classes, while a sufficiently large number of hidden neurons provided support to the output layer during learning. There was also a trend for fewer $n_h / n_o$ needed to reach a performance level of \SI{100}{\%} as the number of output neurons increased. In terms of the time taken to converge in learning, a maximum was found at the first value of $n_h / n_o$ for which the performance first approached \SI{100}{\%}.

The dependence of the performance on the number of target output spikes was examined. In this case, the absolute number of output neurons was fixed at 10, and the number of target output spikes taken over the range: $1 \leq n_s \leq 10$. As before, the network was tasked with classifying 50 input patterns between 10 classes, with five input patterns assigned to each class. 

Fig. \ref{fig12}B shows the minimum fractional number $n_h / n_o$ of hidden to output neurons needed by the network to attain \SI{90}{\%} performance, as a function of the number of target output spikes. An increase in the minimum value of $n_h / n_o$ with the number of target output spikes was found, that showed an indication of levelling off between 8 and 10 output spikes. In terms of the time taken to converge in learning, a small increase of $\sim \SI{25}{\%}$ when learning 10 spikes compared with one spike was measured. When learning a single target output spike, the number of episodes to convergence was measured as \num{5.0 \pm 0.6 e4} in Fig. \ref{fig12}B and close to \num{2.8 \pm 0.2 e4} in Fig. \ref{fig12}A; this apparent discrepancy can be attributed to the more strict criterion used in Fig. \ref{fig12}B, that rejected trials where the performance failed to reach \SI{90}{\%} by the end of learning.

An important consideration when designing any multilayer network is the hidden layer size, and whether it is sufficient to allow for reasonably accurate input classifications during learning. From the above experiments we have quantified the ratio of hidden to output neurons required by the network to allow for accurate classifications to be made, and in particular for the more general case of fully spatio-temporal based output encodings. To the best of our knowledge, this is the first attempt that has aimed to characterise a multilayer network setup for spiking neurons when classifying inputs based on spatio-temporal output patterns.

\subsection*{Biological plausibility}

The technique of backpropagation is commonly associated with poor biological plausibility, an issue that has been challenged in \cite{Gruning2007}. To address this, we propose an alternative and more biologically plausible implementation of our multilayer learning rule.

\paragraph{Reformulation of backpropagation.}

\begin{figure}[t!]
\includegraphics{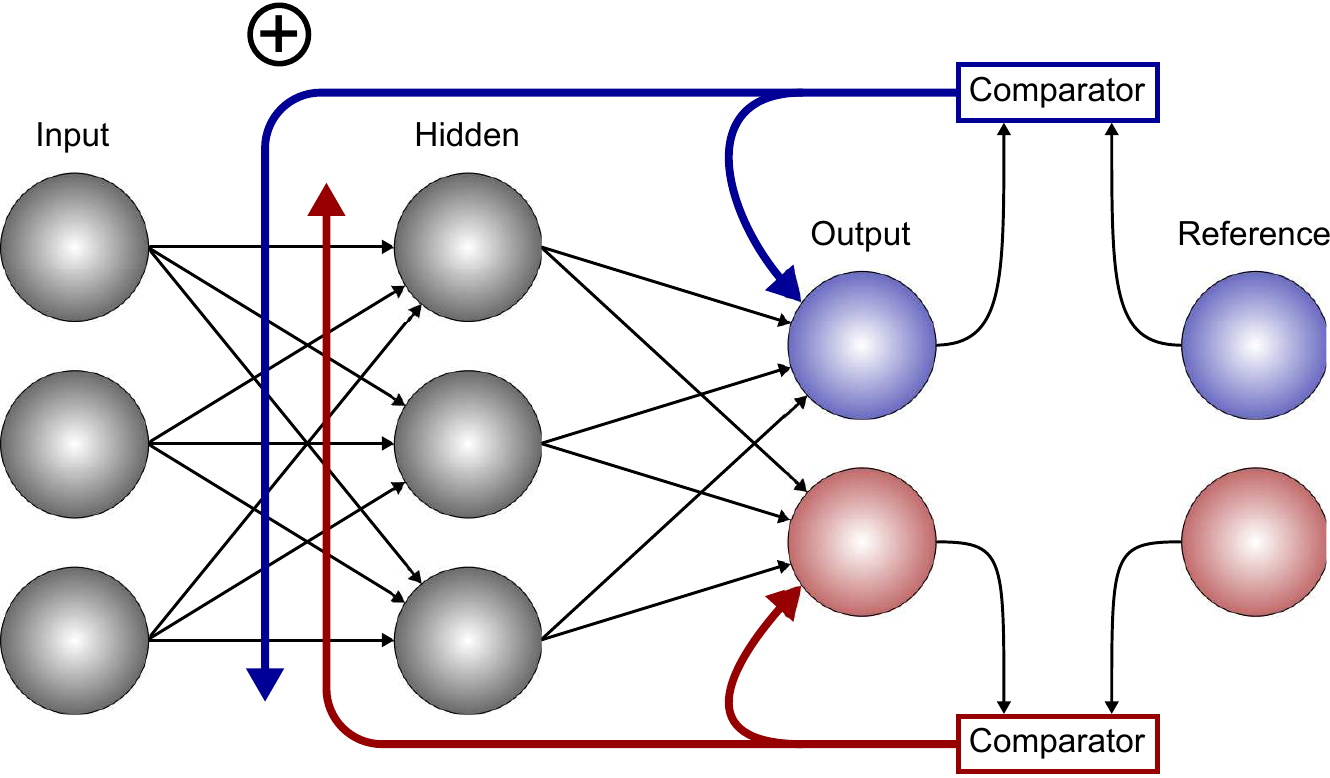}
\caption{
{\bf Biologically plausible implementation of backpropagation in a feedforward multilayer spiking network.}
Input layer neurons are fully connected with hidden layer neurons, which in turn are fully connected with two output neurons. Each output neuron must learn a prescribed target spike train in response to an input pattern presented to the network. It is posited that reference neurons external to the network are available to provide the required target output spike trains during learning. Same colour neurons have a one-one correspondence with each other, such that actual and target activity for each output neuron pair is fed into a unique comparator to provide separate error signals. These error signals are then backpropagated through the network in two stages. At the first stage, each error signal is used to inform the direction and magnitude of synaptic weight changes on its respective output neuron. At the final stage, all error signals are combined together to inform synaptic weight updates on shared hidden neurons.
In this schematic, just two output neurons with corresponding reference outputs are used for clarity, but this setup is similarly extendable to any number of outputs.
}
\label{fig13}
\end{figure}

As the learning rule currently stands, output weight updates (Eq. \ref{eq:output_rule}) can be considered biologically plausible, given that updates have a dependence on locally available pre- and postsynaptic activity variables at each synapse. It is, however, more realistic to effect output weight changes online, for example:
\begin{equation} \label{eq:output_rule_bio}
\dot{w}_{oh}(t) = \eta_o \delta_o(t)\, ( \mathcal{Y}_h \ast \epsilon )(t) \;.
\end{equation}
The supervisory error signals $\delta_o$ are specific to each output neuron, and it is reasonable to suppose that desired postsynaptic activity is provided by an `activity template' external to the network \cite{Knudsen1994}, that is a reference output spike pattern originating in another network. This idea is illustrated in the schematic shown in Fig. \ref{fig13}.

Hidden weight updates (Eq. \ref{eq:hidden_rule}), on the other hand, are more difficult to justify biologically. Although locality at each input-hidden synapse is satisfied in the causal, double convolution term $([ \mathcal{Y}_h( \mathcal{X}_i \ast \epsilon )] \ast \epsilon)$, there is also a non-local dependence on a summation over hidden-specific error signals $\sum_o w_{oh} \delta_o$. It is unclear by which mechanism the strengths of output weights might be communicated back through the network, and further how these weights would then combine with specific error signals to inform synaptic updates.

To provide a biologically plausible reformulation of backpropagated learning, it is therefore necessary to make a few heuristic assumptions regarding the network structure and plasticity. Specifically, we assume output weights are positively valued at initialization and share the same magnitude, and are constrained to positive values during learning. As a result, the dependence of hidden weight updates on the values of output weights can be neglected, thereby providing the modified hidden weight update rule:
\begin{equation} \label{eq:hidden_rule_bio}
\dot{w}_{hi}(t) = 
\frac{\eta_h}{\Delta u_h} 
\sum_o \delta_o(t) ([ \mathcal{Y}_h( \mathcal{X} _i \ast \epsilon )] \ast \epsilon)(t) \;,
\end{equation}
where the summation term $\sum_o \delta_o$ is now a linear combination of output error signals. 
The schematic shown in Fig. \ref{fig13} illustrates this process, and indicates the shared dependence of input-hidden synaptic plasticity changes on multiple output error signals.

We examine the backpropagated error signals $\delta_o$, that are shared between all input-hidden synapses. Biologically, it is plausible that a neuromodulator might perform this function, and particularly given the evidence that neuromodulators can influence both the magnitude and direction of synaptic plasticity changes triggered by \ac{STDP} \cite{Seol2007}. It is further known that the firing activity of dopaminergic neurons can encode a form of error signal \cite{Schultz1997,Schultz2000}, and influence cortiocostriatal plasticity by regulating the concentration of dopamine surrounding each synapse \cite{Reynolds2002}. In light of this, and adopting the method used in \cite{Friedrich2011} as applied to reinforcement learning, previously instantaneous output error signals $\delta_o$ used for hidden and output weight updates are instead substituted for concentration-like variables $\tilde{\delta}_o$, which evolve according to:
\begin{equation} \label{eq:error_filtered}
\tau_D \dot{\tilde{\delta}}_o(t) = -\tilde{\delta}_o(t) + [\mathcal{Z}_o^{\mathrm{ref}}(t) - \mathcal{Z}_o(t)]
\;,
\end{equation}
with a decay time constant $\tau_D = 50$ \si{ms}. In the above, we have also substituted the instantaneous firing rate of an output neuron for its spike train (c.f. Eq. \ref{eq:error_signal}) that is transmittable distally.

\paragraph{Performance of biological backpropagation.}
The performance of the biological implementation of backpropagation (bio-backprop), using Eqs. \ref{eq:output_rule_bio} and \ref{eq:hidden_rule_bio}, was compared against that of the regular multilayer learning rule (backprop). For bio-backprop, output error signals $\delta_o$ were substituted for the filtered signal $\tilde{\delta}_o$ defined in Eq. \ref{eq:error_filtered}. Both learning rules were applied to either a multilayer network containing a single output neuron as the readout, or a multilayer network containing multiple output neurons. In both cases, networks were trained to classify input patterns by the timings of single output spikes.

\begin{figure}[t!]
\includegraphics{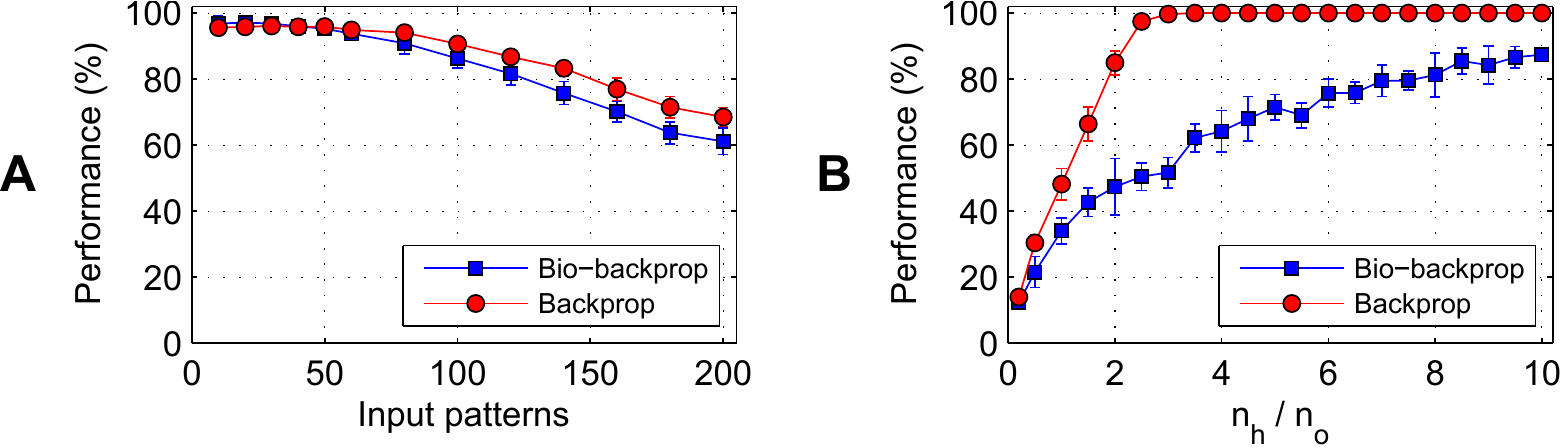}
\caption{
{\bf Performance of a biologically plausible implementation of backpropagation.}
A biologically plausible implementation of backpropagation (bio-backprop) was used to drive hidden synaptic weight updates. In both (A) and (B), the network was trained to classify input patterns equally assigned between 10 classes by the timing of single output spikes.
(A) The performance of bio-backprop as a function of the number of input patterns presented to the network, based on the experiment in Fig. \ref{fig8}. The network contained 10 hidden neurons and a single output neuron. Results were averaged over 20 independent runs.
(B) The performance of bio-backprop as a function of the ratio of hidden to output neurons, based on the experiment in Fig. \ref{fig12}. The network contained an increasing number $n_h$ of hidden neurons, and $n_o = 10$ output neurons. A fixed number of input patterns $p = 50$ was used. For reference, results from the regular backpropagation learning rule on the same learning task (labelled Backprop) is included (Fig. \ref{fig12}). Results were averaged over 10 independent runs.
}
\label{fig14}
\end{figure}

The single-output network contained 10 hidden neurons, and was tasked with classifying an increasing number of input patterns into 10 classes (c.f. experiment of Fig. \ref{fig8}). As shown in Fig. \ref{fig14}A, little difference was found in the performance between the bio-backprop and backprop learning rules for less than 80 input patterns. However, as the number of input patterns increased there was a small performance difference in favour of backprop, approaching \SI{8}{\%} by 200 input patterns. In terms of the convergence time, bio-backprop was consistently slower than backprop, taking at least 1.5 times the number of episodes needed by backprop to complete learning.

We next considered a multiple-output network containing $n_o = 10$ output neurons and an increasing number $n_h$ of hidden neurons. The network was tasked with classifying 50 input patterns into 10 classes (c.f. experiment of Fig. \ref{fig12}). From Fig. \ref{fig14}B, a marked difference in the performance favouring backprop over bio-backprop was apparent; while it took just two times the number of hidden to output neurons for backprop to reach a \SI{90}{\%} performance level, it took bio-backprop at least ten times the number of hidden to output neurons to reach the same performance level. In other words, bio-backprop needed five times the number of hidden neurons as backprop to accurately classify all 50 input patterns presented to the network. With respect to the convergence time, bio-backprop took almost 1.5 times as many episodes as backprop to reach a \SI{90}{\%} performance level, using $n_h / n_o = 10$ and $n_h / n_o = 2$ for bio-backprop and backprop respectively.

From these results, it is clear that the heuristic bio-backprop rule can maintain a similar level of performance to the analytical backprop rule for networks containing a single output neuron. However, for networks containing more than one output neuron the performance of bio-backprop lagged behind that of backprop; a reasonable performance level for bio-backprop could only be recovered by a large increase in the number of hidden neurons. Furthermore, weight distributions (not shown) indicated a reduced ability of networks trained with bio-backprop to effectively distribute the synaptic load between hidden layer neurons: individual hidden neurons either contributed intensely or weakly to the activity over all output neurons. Ideally, individual hidden neurons would instead be selected to contribute to specific output responses, which was evidenced for the backprop rule (c.f. Fig. \ref{fig11}C).

Despite some limitations, bio-backprop still proved to be a capable learning rule, and maintained a performance level well above the chance level of \SI{10}{\%} in most cases. As such, bio-backprop represents an alternative to our analytical backprop rule when increased biological plausibility is desired.

\section*{Discussion}
In this paper we have presented a multilayer learning rule for a spiking network containing a hidden layer that combines gradient ascent with the technique of backpropagation, and demonstrated its high performance when encoding a large number of input patterns by the precise timings of multiple output spikes. We have further provided an alternative and more biologically plausible implementation of our learning rule, and discussed the underlying neural mechanisms which might support a form of backpropagated learning in the nervous system. Our approach complements the recurrent network algorithms proposed in \cite{Brea2011,Brea2013,JimenezRezende2014}.

In our analysis we used the escape noise neuron model defined in \cite{Gerstner2002}, which has been shown to closely approximate the variable firing activity of neurons \emph{in-vivo} \cite{Jolivet2006}. Our choice of neuron model was primarily motivated by its general applicability in a wide range of learning paradigms, including supervised \cite{Pfister2006,Brea2013} and reinforcement \cite{Florian2007,Fremaux2010} learning. A key advantage of implementing escape noise neurons comes from being able to determine the probability of generating a specific output spike pattern \cite{Pfister2006}, which can then form the basis of a suitable objective function. Here we took the approach of maximizing the log-likelihood of generating a desired output spike pattern in a multilayer network through a combination of gradient ascent and backpropagation, that is an extension of the single-layer learning rule proposed by \cite{Pfister2006} to multilayer networks. Output weight updates result from a product of locally available pre- and postsynaptic activity terms, that bears a resemblance to Hebbian-like learning; the presynaptic term originates from filtered hidden neuron spike trains as a \ac{PSP}, and the postsynaptic term an output error signal that guides the direction and magnitude of weight changes. Hidden weight updates, however, appear as a three-factor rule: \ac{PSP}s due to input spikes are combined with hidden spike trains, to then be modulated by a linear combination of backpropagated error signals to allow for hidden weight changes.

From training multilayer networks to map between input-output spiking patterns, it proved necessary to represent input patterns with a sufficiently high degree of spiking activity at each input neuron; sparse representations otherwise led to decreased performance. This requirement is apparent from an examination of the hidden layer weight update rule, which has a dependence on hidden neuron spike trains: a lack of input-driven hidden layer activity prevented weight updates from taking place, thereby resulting in diminished learning. Previous multilayer learning rules \cite{Bohte2004,Sporea2013} have faced a similar challenge in effectively presenting input patterns to the network, but instead took the approach of introducing multiple synaptic connections with varying conduction delays between neurons of neighbouring layers: also termed subconnections. For example, one presynaptic spike would evoke multiple \ac{PSP}s at each of its postsynaptic targets through the many subconnections available, thereby driving a sufficient level of hidden layer activity. Combining sparsely encoded input patterns with multiple subconnections represents a plausible alternative to the method we employed in this paper, and could be advantageous when applied to real world datasets such as Iris, for which sparse representations of inputs can be ideal \cite{Sporea2013}.

We were motivated to introduce synaptic scaling to the network to maintain an optimal range of hidden firing rates \cite{VanRossum2000}, a process that has been observed in biological networks \cite{Turrigiano1998}. Aside from effecting the firing rate, the introduction of synaptic scaling also has side benefits: such as removing the networks over-reliance on the initial values of synaptic weights \cite{Sporea2013}, a critical issue that was identified in \cite{Bohte2004}.

An important contribution of our paper is the large number of pattern encodings that can be performed by our learning rule: in comparison with multilayer \ac{ReSuMe} \cite{Sporea2013}, trained on a similar network setup, our learning rule was capable of at least $10 \times$ as many pattern encodings at a \SI{90}{\%} performance level (Fig. \ref{fig8}). Furthermore, we believe our encoding method better took advantage of spike-timing than most alternative methods \cite{Gutig2006,Florian2012,Mohemmed2012}; for example, the Tempotron \cite{Gutig2006} can only classify input patterns into two classes using a spike / no-spike coding scheme, and the experiments run for the Chronotron \cite{Florian2012} and SPAN \cite{Mohemmed2012} required precisely matched output-target spike pairs, which could result in overlearning and impact negatively on generalizing to new input patterns. In particular, we found our encoding method allowed for increased performance when using multiple target output spikes, and especially for a larger number of hidden layer neurons. Multiple output spikes also allowed the network to better generalize to new input patterns.

Backpropagated learning is commonly considered to lack biological plausibility for two key reasons: the first being mechanistic, in the sense that synapses are used bidirectionally to communicate activity variables both forwards and backwards throughout the network; the second reason is cognitively, since a fully specified target signal must be provided to the network during learning \cite{Gruning2007}.

In this paper we have addressed the first implausibility, by reformulating our hidden layer weight update rule to depend on just local input-hidden synaptic activity variables and a global, linear combination of output-specific error signals (Fig. \ref{fig13}). Hence, our interpretation relies on the presence of error signals that are diffusely available across a large number of synapse. Biologically, this role might be fulfilled by a neuromodulator such as dopamine, which is known to act as an error-correcting signal by modulating synaptic plasticity changes in the cortico-striatal junction during behavioural learning tasks \cite{Reynolds2002}.

Like most existing learning rules for spiking networks we have assumed the presence of a supervisory signal, which is used to perform on-line comparisons between actual and target output spike patterns during learning. Although this might be deemed cognitively implausible, it is possible that such a signal originates from an external network that acts as an `activity template' \cite{Knudsen1994}, which can explain functional plasticity changes in neurons encoding for auditory stimuli in the barn owl \cite{Knudsen2002}. However, more recently, reward-modulated learning rules for spiking networks have emerged as a more plausible alternative to supervised learning \cite{Florian2007,Izhikevich2007}, which instead provide summary feedback on the correctness of network responses. It has been shown in \cite{Farries2007,Fremaux2010} and in our previous work \cite{Gardner2013} how reward-modulation can be applied to learning precise output spiking patterns.

\paragraph{Conclusions and future work.} In principle, our multilayer learning rule follows from those previously introduced in \cite{Bohte2004,Sporea2013} which have adapted backpropagaton for use in feedforward spiking networks. Through several benchmark tests in this paper we have indicated the high performance of our learning rule, thereby lending support to its practical deployment as an efficient neural classifier. We have further highlighted the advantages of using a fully temporal code based on multiple output spike-timings to reliably encode for input patterns. Finally, to address the biological shortcomings of backpropagated learning, we presented a heuristic reformulation of our learning rule which we argue can be considered biologically plausible. 

The framework in which we have developed our learning rule is general, and has found applications in the areas of both supervised and reinforcement learning for feedforward and recurrent network structures. It is therefore natural to assume that our supervised rule might have a reinforcement analogue, that instead uses a delayed feedback signal to indicate the overall `correctness' of network responses during learning. In Grüning \cite{Gruning2007}, it has been shown how backpropagation can be reimplemented as a cognitively more plausible reinforcement learning scheme, but for rate-coded neurons; future work could attempt to relate such a technique to our own rule for spiking neurons, with the intent of supporting a biological backpropagation rule.

\section*{Methods}
\subsection*{Neuron model}

We start by considering a single postsynaptic neuron $o$ in the network that receives input from $1 \leq h \leq n_h$ presynaptic neurons. The list of presynaptic spikes due to neuron $h$, up to time $t$, is $y_h(t) = \{t_h^1,...,\hat{t}_h < t\}$, where $\hat{t}_h$ is always the last spike before $t$. If the postsynaptic neuron $o$ generates the list of output spikes $z_o(t) = \{t_o^1,...,\hat{t}_o < t\}$ in response to the presynaptic pattern $y_h \in \mathbf{y}$, then its membrane potential at time $t$ is defined by the \acf{SRM} \cite{Gerstner2002}:
\begin{equation} \label{eq:potential}
u_o(t|\mathbf{y}, z_o) := \sum_h w_{oh} \int_0^t \mathcal{Y}_h(t') \epsilon(t-t') \mathrm{d}t' + \int_0^t \mathcal{Z}_o(t') \kappa(t-t') \mathrm{d}t' \;, 
\end{equation}
where $w_{oh}$ is the synaptic weight between neurons $o$ and $h$, and $\mathcal{Y}_h(t)$ and $\mathcal{Z}_o(t)$ are the presynaptic and postsynaptic spike trains respectively, with a spike train defined in terms of a sum of Dirac $\delta$ functions: $\mathcal{Y}_h(t)=\sum_f\delta(t-t_h^f)$. $\epsilon(s)$ and $\kappa(s)$ are the \ac{PSP} and reset kernels respectively, taken as:
\begin{equation} \label{eq:kernels}
\epsilon(s) = \epsilon_0\, [\mathrm{e}^{-s / \tau_m} - \mathrm{e}^{-s / \tau_s} ]\, \Theta(s)\;\;\; \mathrm{and}\;\;\; \kappa(s) = \kappa_0 \mathrm{e}^{-s/ \tau_m}\, \Theta(s) \;,
\end{equation}
where $\epsilon_0=4$ \si{mV} and $\kappa_0=-15$ \si{mV} are scaling constants, $\tau_m=10$ \si{ms} the membrane time constant, $\tau_s=5$ \si{ms} the synaptic rise time and $\Theta(s)$ the Heaviside step function.

Neuronal spike events are generated by a point process with stochastic intensity $\rho_o(t)$, that is the instantaneous firing rate of a postsynaptic neuron, where the probability of generating a spike at time $t$ over a small time interval $[t, t + \delta t)$ is given by $\rho_o(t)\delta t$. The firing rate has a nonlinear dependence on the postsynaptic neuron's membrane potential, that in turn depends on both its presynaptic input and the postsynaptic neuron's firing history: $\rho_o(t|\mathbf{y}, z_o) = g[u_o(t|\mathbf{y}, z_o)]$. Here, we take an exponential dependence of the firing rate on the distance between the membrane potential and firing threshold $\vartheta$ \cite{Gerstner2002}:
\begin{equation} \label{eq:EXP_rate}
g[u] = \rho_0 \exp \left( \frac{ u-\vartheta }{ \Delta u } \right) \;,
\end{equation}
with the instantaneous firing rate at threshold $\rho_0=0.01$ \si{ms^{-1}} and $\vartheta=15$ \si{mV}. The smoothness of the threshold was set to $\Delta u_o=0.2$ \si{mV} for output layer neurons and $\Delta u_h=2$ \si{mV} for hidden layer neurons. In the limit $\Delta u \rightarrow 0$ the deterministic \ac{LIF} model can be recovered \cite{Gerstner2002}. Our choice of $\Delta u_h > \Delta u_o$ was motivated by the need for increased variation in hidden neuron spiking for learning to succeed, as indicated by preliminary results. Taking an exponential dependence of the firing rate on the membrane potential represents one choice for distributing output spikes; alternative functional dependencies exist, such as the Arrhenius \& Current model \cite{Plesser2000}, which we have previously applied to learning temporally precise spiking patterns in \cite{Gardner2013}.

\subsection*{Learning rule}

We initially derive weight update rules for the connections between the hidden and output layers, as originally shown by Pfister et al. \cite{Pfister2006}. We then extend our analysis to include weight updates between the input and hidden layers using backpropagation, that is our novel contribution of a multilayer learning rule in a network of spiking neurons. In our notation, input layer neurons are indexed as $i$, hidden neurons $h$ and output neurons $o$.

\paragraph{Objective function.}
Implementing stochastic spiking neurons allows us to determine the likelihood of generating a prescribed target spike train.
Specifically, the probability density of an output neuron $o$ generating a list of target output spikes $z_o^{\mathrm{ref}} = \{\tilde{t}_o^1,\tilde{t}_o^2,...\}$ in response to a hidden spike pattern $\mathbf{y}$ is given by \cite{Pfister2006}:
\begin{equation} \label{eq:PDF_single}
P( z_o^{\mathrm{ref}}|\mathbf{y}) = \exp \bigg( \int_0^T \log \left( \rho_o(t|\mathbf{y}, z_o) \right) \mathcal{Z}_o^{\mathrm{ref}}(t) - \rho_o(t|\mathbf{y}, z_o) \mathrm{d}t \bigg) \;,
\end{equation}
where $\mathcal{Z}_o^{\mathrm{ref}}(t)=\sum_f\delta(t-\tilde{t}_o^f)$ and $T$ is the duration over which $\mathbf{y}$ is presented.
It is noted that output neuron activity implicitly depends on variable hidden layer activity through the functional dependence $\rho_o(t) = g[u_o(t)]$ (c.f. Eqs. \ref{eq:potential} and \ref{eq:EXP_rate}). For more than one output neuron, the probability density of generating a spatio-temporal target output pattern $z_o^{\mathrm{ref}} \in \mathbf{z}^{\mathrm{ref}}$ is given by
\begin{align} \label{eq:PDF_pattern}
P( \mathbf{z}^{\mathrm{ref}}|\mathbf{y} ) &= \prod_o P( z_o^{\mathrm{ref}}|\mathbf{y} ) \nonumber \\ 
																			&= \exp \bigg( \sum_o \int_0^T  \log \left( \rho_o(t|\mathbf{y}, z_o) \right) \mathcal{Z}_o^{\mathrm{ref}}(t) - \rho_o(t|\mathbf{y}, z_o) \mathrm{d}t \bigg) \;.
\end{align}
Taking the logarithm of Eq. \ref{eq:PDF_pattern} provides us with an objective function, that is a smooth function of the network parameters:
\begin{equation} \label{eq:log_PDF_pattern}
\log P( \mathbf{z}^{\mathrm{ref}}|\mathbf{y} ) = \sum_o \int_0^T \log \left( \rho_o(t|\mathbf{y}, z_o) \right) \mathcal{Z}_o^{\mathrm{ref}}(t) - \rho_o(t|\mathbf{y}, z_o) \mathrm{d}t \;.
\end{equation}
By gradient ascent we seek to optimize the above function, through adjusting plastic parameters in the network. Here, we focus on changing the values of synaptic weights, although other network parameters such as conduction delays might also be trained.

\paragraph{Output weight updates.}
Taking the positive gradient of the log-likelihood (Eq. \ref{eq:log_PDF_pattern}) provides us with the direction of weight updates for neurons in the output layer, such that the expectation of generating a target output pattern $\mathbf{z}^{\mathrm{ref}}$ is increased, i.e.:
\begin{equation} \label{eq:w_oh_update1}
\Delta w_{oh} = \eta_o \frac{\partial \log P( \mathbf{z}^{\mathrm{ref}}|\mathbf{y} )}{\partial w_{oh}} \;,
\end{equation}
where $\eta_o$ is the output layer learning rate. The derivative of the log-likelihood can be found as
\begin{equation} \label{eq:Do_log_PDF_pattern}
\frac{\partial \log P( \mathbf{z}^{\mathrm{ref}}|\mathbf{y} )}{\partial w_{oh}}  = 
\int_0^T 
\frac{ \rho_o'(t|\mathbf{y}, z_o) }{ \rho_o(t|\mathbf{y}, z_o) } \left[ \mathcal{Z}_o^{\mathrm{ref}}(t) - \rho_o(t|\mathbf{y}, z_o) \right] ( \mathcal{Y}_h \ast \epsilon )(t) \mathrm{d}t \;,
\end{equation}
where $\rho_o'(t|\mathbf{y}, z_o) = \frac{\mathrm{d} g(u)}{\mathrm{d} u}|_{u = u_o(t|\mathbf{y}, z_o)}$, and $( \mathcal{Y}_h \ast \epsilon )(t)$ is the convolution of the hidden spike train $\mathcal{Y}_h(t')$ with the \ac{PSP} kernel $\epsilon(t-t')$ that is defined in Eq. \ref{eq:convolution}.
Given our choice of an exponential dependence for the firing rate on the membrane potential, defined in Eq. \ref{eq:EXP_rate}, it follows that
\begin{equation} \label{eq:D_frac_EXP_rate}
\frac{ \rho_o'(t|\mathbf{y}, z_o) }{ \rho_o(t|\mathbf{y}, z_o) } = \frac{1}{\Delta u_o} \;,
\end{equation}
hence combining Eqs. \ref{eq:w_oh_update1}, \ref{eq:Do_log_PDF_pattern} and \ref{eq:D_frac_EXP_rate} provides the output layer weight update rule:
\begin{equation} \label{eq:w_oh_update2}
\Delta w_{oh} = \frac{\eta_o}{\Delta u_o} \int_0^T \left[ \mathcal{Z}_o^{\mathrm{ref}}(t) - \rho_o(t|\mathbf{y}, z_o) \right]\, ( \mathcal{Y}_h \ast \epsilon )(t) \mathrm{d}t \;.
\end{equation}
We define the backpropagated error signal $\delta_o$ for the $o^{\mathrm{th}}$ output neuron as
\begin{equation} \label{eq:backprop_error}
\delta_{o}(t|\mathbf{y}, z_o) := \frac{1}{\Delta u_o} \left[ \mathcal{Z}_o^{\mathrm{ref}}(t) - \rho_o(t|\mathbf{y}, z_o) \right] \;,
\end{equation}
that is substituted into Eq. \ref{eq:w_oh_update2} for compactness:
\begin{equation} \label{eq:w_oh_update}
\Delta w_{oh} = \eta_o \int_0^T \delta_o(t|\mathbf{y}, z_o)\, ( \mathcal{Y}_h \ast \epsilon )(t)\, \mathrm{d}t \;.
\end{equation}
The above supervised learning rule was originally derived by Pfister et al. \cite{Pfister2006} for a single-layer network, that optimizes the log-likelihood of generating a desired postsynaptic spike train in response to a given input pattern.

\paragraph{Hidden weight updates.}
Continuing through to the hidden layer, weights between input and hidden layer neurons are updated according to
\begin{equation} \label{eq:Dh_log_PDF_pattern}
\Delta w_{hi} = 
\eta_h \frac{\partial \log P( \mathbf{z}^{\mathrm{ref}}|\mathbf{y} )}{\partial w_{hi}}
\; ,
\end{equation}
where $\eta_h$ is the hidden layer learning rate. Using Eq. \ref{eq:log_PDF_pattern}, and by making use of the chain rule, the gradient of the log-likelihood with respect to hidden layer weights can be expressed as
\begin{align} \label{eq:Dh_log_PDF_pattern1}
\frac{\partial \log P( \mathbf{z}^{\mathrm{ref}}|\mathbf{y} )}{\partial w_{hi}} 
&=
\sum_o \int_0^T \frac{\partial}{\partial w_{hi}} \left[ \log \left( \rho_o(t|\mathbf{y}, z_o) \right) \mathcal{Z}_o^{\mathrm{ref}}(t) - \rho_o(t|\mathbf{y}, z_o) \right] \mathrm{d}t \nonumber \\
&=
\sum_o \int_0^T \frac{ \rho_o'(t|\mathbf{y}, z_o) }{ \rho_o(t|\mathbf{y}, z_o) } \left[ \mathcal{Z}_o^{\mathrm{ref}}(t) - \rho_o(t|\mathbf{y}, z_o) \right] \frac{\partial u_o(t|\mathbf{y}, z_o)}{\partial w_{hi}} \mathrm{d}t
\; .
\end{align}
Using Eqs. \ref{eq:D_frac_EXP_rate} and \ref{eq:backprop_error}, the above can be compacted:
\begin{equation} \label{eq:Dh_log_PDF_pattern2}
\frac{\partial \log P( \mathbf{z}^{\mathrm{ref}}|\mathbf{y} )}{\partial w_{hi}} =
\sum_o \int_0^T \delta_o(t|\mathbf{y}, z_o) \frac{\partial u_o(t|\mathbf{y}, z_o)}{\partial w_{hi}} \mathrm{d}t
\; .
\end{equation}
The membrane potential of an output layer neuron has a dependence on the firing activity of neurons in the hidden layer according to Eq. \ref{eq:potential}, hence the second term on the right-hand side of Eq. \ref{eq:Dh_log_PDF_pattern2} can be rewritten as
\begin{equation} \label{eq:Dh_log_PDF_pattern3}
\frac{\partial u_o(t|\mathbf{y}, z_o)}{\partial w_{hi}} = w_{oh}\; \frac{\partial}{\partial w_{hi}} (\mathcal{Y}_h \ast \epsilon)(t) \;.
\end{equation}
Weights changes take place on a time scale of $T \gg \tau_m$, therefore the gradient of the convolution $( \mathcal{Y}_h \ast \epsilon )(t)$ can be well approximated by
\begin{equation} \label{eq:Dh_log_PDF_pattern4}
\frac{\partial}{\partial w_{hi}} ( \mathcal{Y}_h \ast \epsilon )(t) \approx \int_0^t \frac{\partial \mathcal{Y}_h(t')}{\partial w_{hi}} \epsilon(t-t') \mathrm{d}t' \;.
\end{equation}
The spike train $\mathcal{Y}_h(t')$ is a discontinuous random variable with no smooth dependence on network parameters, leaving the gradient $\frac{\partial \mathcal{Y}_h(t')}{\partial w_{hi}}$ difficult to solve analytically. Therefore, applying the technique used in \cite{Fremaux2013}, we heuristically make the substitution $\mathcal{Y}_h(t') \rightarrow \left\langle \mathcal{Y}_h(t')\right\rangle_{y_h|\mathbf{x}}$, that is the expectation of the hidden spike train $\mathcal{Y}_h(t')$ conditioned on the input pattern $\mathbf{x}$. The expectation of $\mathcal{Y}_h(t')$ has a smooth dependence on network parameters, and its gradient is given by:
\begin{align} \label{eq:Dh_log_PDF_pattern5}
\frac{\partial \left\langle \mathcal{Y}_h(t') \right\rangle_{y_h|\mathbf{x}}}{\partial w_{hi}} 
&=
\frac{\partial}{\partial w_{hi}} \int \mathcal{Q}(t') P(y_h = q|\mathbf{x}) \mathrm{d}q \nonumber \\
&=
\int \mathcal{S}(t') P(y_h = q|\mathbf{x}) \frac{\partial \log P(y_h = q|\mathbf{x})}{\partial w_{hi}} \mathrm{d}q \;,
\end{align}
where we have used the relation $\frac{1}{P}\frac{\partial P}{\partial w_{hi}} = \frac{\partial \log P}{\partial w_{hi}}$, the integral runs over all possible lists of spikes $q(t') = \{t^1, t^2, ..., \hat{t} < t'\}$ up to time $t'$ and $\mathcal{Q}(t') = \sum_{t^f \in q} \delta(t' - t^f)$ is a spike train. $P(y_h=q|\mathbf{x})$ is the probability density of the list of hidden spikes $y_h$ being equal to $q$, conditioned on $\mathbf{x}$. The probability density or likelihood of a hidden neuron generating a list of spikes $q$ up to $t'$ in response to $\mathbf{x}$ is defined similarly to equation \ref{eq:PDF_single}:
\begin{equation} \label{eq:PDF_hidden}
P( y_h = q|\mathbf{x} ) = \exp \bigg( \int_0^{t'} \log \left( \rho_h(s|\mathbf{x}, q) \right) \mathcal{Q}(s) - \rho_h(s|\mathbf{x}, q) \, \mathrm{d}s \bigg) \;,
\end{equation}
and the gradient of the log-likelihood:
\begin{equation} \label{eq:Dh_log_PDF_pattern6}
\frac{\partial \log P( y_h = q|\mathbf{x} )}{\partial w_{hi}}  = 
\frac{1}{\Delta u_h} \int_0^{t'} 
\left[ \mathcal{Q}(s) - \rho_h(s|\mathbf{x}, q) \right] ( \mathcal{X}_i \ast \epsilon )(t)\, 
\mathrm{d}s \;.
\end{equation}
Hence, Eq. \ref{eq:Dh_log_PDF_pattern5} becomes:
\begin{equation} \label{eq:Dh_log_PDF_pattern7}
\frac{\partial \left\langle \mathcal{Y}_h(t') \right\rangle_{y_h|\mathbf{x}}}{\partial w_{hi}} 
=
\frac{1}{\Delta u_h} \int \mathcal{S}(t') P(y_h = q|\mathbf{x}) \left( \int_0^{t'} 
\left[ \mathcal{Q}(s) - \rho_h(s|\mathbf{x}, q) \right] ( \mathcal{X}_i \ast \epsilon )(t)\, 
\mathrm{d}s \right) \mathrm{d}q \;,
\end{equation}
such that a spike generated by the neuron at time $t'$ depends not only on recent input spikes, but also on its own entire spiking history $q$ through the integration between times $0$ and $t'$.

The above can be simplified if we choose to neglect the neuron's firing history by taking the last hidden spike time $\hat{t}'_h < t'$ as given, that allows the substitution $\mathcal{Y}_h(t') \rightarrow \left\langle \mathcal{Y}_h(t')\right\rangle_{y_h|\mathbf{x},\hat{t}'_h}$ for the expectation of the hidden spike train $\mathcal{Y}_h(t')$ conditioned on both the input $\mathbf{x}$ and $\hat{t}'_h$ \cite{Fremaux2013}. In this case, neglecting the neuron's firing history is not an unreasonable choice, given that the gradient of $\mathcal{Y}_h(t')$ is convolved by the exponential \ac{PSP} kernel $\epsilon$ in Eq. \ref{eq:Dh_log_PDF_pattern4} that already captures the recent firing history of the neuron. Hence, the gradient of the hidden spike train in Eq. \ref{eq:Dh_log_PDF_pattern4} can instead be expressed in terms of the value of a spike train $\mathcal{Q}$ at each point in time:
\begin{align} \label{eq:Dh_log_PDF_pattern8}
\frac{\partial \left\langle \mathcal{Y}_h(t') \right\rangle_{y_h|\mathbf{x},\hat{t}_h'}}{\partial w_{hi}} 
&= 
\frac{\partial}{\partial w_{hi}} \sum_{\mathcal{Q} \in \{0,\delta(t')\}} \mathcal{Q}(t') P(y_h=q|\mathbf{x},\hat{t}_h') \nonumber \\
&= 
\delta(t'-\hat{t}') \frac{\partial \rho_h (t'|\mathbf{x},\hat{t}_h')}{\partial w_{hi}} 
\;,
\end{align}
where we have used the identity $P(y_h=q|\mathbf{x},\hat{t}_h') = \rho_h (t'|\mathbf{x},\hat{t}_h')$ and $\delta(t'-\hat{t}')$ is the Dirac distribution as a function of a last spike $\hat{t}'$. Using Eqs. \ref{eq:potential} and \ref{eq:EXP_rate} we find:
\begin{align} \label{eq:Dh_log_PDF_pattern9}
\frac{\partial \left\langle \mathcal{Y}_h(t') \right\rangle_{y_h|\mathbf{x},\hat{t}_h'}}{\partial w_{hi}}  
&= 
\frac{1}{\Delta u_h} \delta(t'-\hat{t}') \rho_h (t'|\mathbf{x},\hat{t}_h') ( \mathcal{X}_i \ast \epsilon )(t') \nonumber \\
&=
\frac{1}{\Delta u_h} \left\langle \mathcal{Y}_h ( \mathcal{X}_i \ast \epsilon ) \right\rangle_{y_h|\mathbf{x},\hat{t}_h'} \;.
\end{align}
On each learning episode, our best estimate for the expected gradient comes from considering the current observation of $y_h$ given $\mathbf{x}$; hence, the expectation can be dropped and the above can be combined with Eqs. \ref{eq:Dh_log_PDF_pattern3} and \ref{eq:Dh_log_PDF_pattern4} to give
\begin{equation} \label{eq:Dh_log_PDF_pattern10}
\frac{\partial u_o(t|\mathbf{y}, z_o)}{\partial w_{hi}} = \frac{w_{oh}}{\Delta u_h} ([ \mathcal{Y}_h( \mathcal{X}_i \ast \epsilon )] \ast \epsilon)(t) \;,
\end{equation}
where we have defined a double convolution as:
\begin{equation} \label{eq:double_convolution}
([ \mathcal{Y}_h( \mathcal{X}_i \ast \epsilon )] \ast \epsilon)(t) \equiv \int_0^t \mathcal{Y}_h(t') \left[ \int_0^{t'} \mathcal{X}_i(t'') \epsilon(t'-t'') \mathrm{d}t'' \right] \epsilon(t-t') \mathrm{d}t' \;.
\end{equation}
Finally, combining Eq. \ref{eq:Dh_log_PDF_pattern10} with Eqs. \ref{eq:Dh_log_PDF_pattern} and \ref{eq:Dh_log_PDF_pattern2} provides the hidden layer weight update rule:
\begin{equation} \label{eq:w_hi_update}
\Delta w_{hi} = 
\frac{\eta_h}{\Delta u_h} 
\sum_o w_{oh} \int_0^T
\delta_o(t|\mathbf{y}, z_o) ([ \mathcal{Y}_h( \mathcal{X}_i \ast \epsilon )] \ast \epsilon)(t) \mathrm{d}t \;.
\end{equation}

\subsection*{Synaptic scaling}

For hidden layer weight updates to take place, a degree of background hidden neuron spiking is necessary during learning. This condition can be satisfied if we apply synaptic scaling to hidden layer weights, that has previously been shown to maintain a homeostatic firing rate and introduce competition between afferent connections \cite{VanRossum2000}. Therefore, in addition to Eq. \ref{eq:w_hi_update}, hidden weights are modified at the end of each learning episode by the scaling rule:
\begin{equation} \label{eq:scaling}
 \Delta w_{hi} = 
  \begin{cases}
   \gamma \; |w_{hi}| \; (\nu_{\mathrm{max}} - \nu_h) & \text{if } \nu_h > \nu_{\mathrm{max}} \\
   \gamma \; |w_{hi}| \; (\nu_{\mathrm{min}} - \nu_h) & \text{if } \nu_h < \nu_{\mathrm{min}} \;,
  \end{cases}
\end{equation}
where $\gamma = \num{1e-2}$ is the scaling strength, $\nu_h$ the actual firing rate of the $h^{\mathrm{th}}$ hidden neuron and $\nu_{\mathrm{max}} = \SI{40}{Hz}$ and $\nu_{\mathrm{max}} = \SI{2}{Hz}$ the maximum and minimum reference firing rates respectively. The above drives the firing rate of each hidden neuron to remain within the range $2 \leq \nu_h \leq 40$ \si{Hz}, that makes the network less sensitive to its initial state and prevents extremes in the firing activity of hidden neurons \cite{Sporea2013}.

\subsection*{Pattern statistics}

Input patterns were presented to the network by $n_i = 100$ input layer neurons, where each input neuron contained an independent Poisson spike train with a mean firing rate of \SI{6}{Hz}. A relative refractory period with a time constant of \SI{10}{ms} was simulated when generating each spike train for biological realism. A random realization of each input pattern was used, for a total of $p$ different patterns. 

Learning took place on an episodic basis, where each episode corresponded to the presentation of an input pattern to the network lasting duration $T = \SI{500}{ms}$. The order in which input patterns were presented was random. Unless otherwise stated, simulations were run over $1000 p$ episodes to ensure a sufficient amount of time for the network to learn the desired number of inputs. Hence, on average, each input pattern was presented 1000 times.

Every input pattern was associated with a target output pattern, and multiple inputs belonging to the same class shared the same target output. A target output pattern consisted of a predetermined spike train at each output neuron, and target spike trains contained the same number of spikes $n_s \in \{1, 10\}$ at each output that depended on the learning task. Target spike trains were initialized by randomly selecting each target spike time $\tilde{t}^f$ from a uniform distribution over the interval $\tilde{t}^f \in \left[ 40, T \right) \si{ms}$, with an interspike separation of at least \SI{10}{ms} to avoid conflicted output responses during learning. A minimum target spike-timing of 40 \si{ms} was taken given the evidence that values $\tilde{t}^f < 4 \tau_m$ led to reduced performance \cite{Florian2012}. 

At each output neuron, target spike trains differed from each other by a minimum distance of $\mathcal{D}_{\mathrm{min}} > n_s / 2$ to ensure each input class was assigned a unique target response, and to reduce crosstalk during learning. The minimum distance scaled with the number of target output spikes, which increased the separation between classes. For our definition of $\mathcal{D}_{\mathrm{min}}$ and choice of $T$, a maximum of $c = 66$ classes identified by a single target output spike was supported, and more correspondingly for multiple target output spikes.

\subsection*{Pattern recognition}

Networks were trained to classify input patterns by the timing of output spikes, such that multiple inputs belonging to the same class shared the same target output. Target outputs were randomly set at the start of each simulation, and networks were trained to assign $p$ input patterns between $c$ classes. For each class, a target output contained between one and ten spikes, depending on the learning task.

Instead of relying on precisely matched actual and target output spike trains to classify inputs, we instead allowed for sufficiently accurate output spike trains, that were closer to their desired targets in comparison with any other potential target. In discriminating between different classes of input patterns we used the \acf{vRD} \cite{Rossum2001}, that is a metric for the temporal distance between two spike trains. 

From considering a list of spikes $t_f \in z$, the \ac{vRD} is computed by first performing a convolution over a spike train $\mathcal{Z}(t) = \sum_f \delta (t - t^f)$ with an exponential function:
\begin{equation} \label{eq:vRD0}
\tilde{\mathcal{Z}}(t) = \sum_f \exp \left( - \frac{t - t^f}{\tau_c} \right) \Theta(t - t^f) \;,
\end{equation}
where we set the coincidence time constant $\tau_c = 10$ \si{ms}. Hence, using the above, we can obtain $\tilde{\mathcal{Z}}_o$ and $\tilde{\mathcal{Z}}_o^{\mathrm{ref}}$ from $\mathcal{\mathcal{Z}}_o$ and $\mathcal{\mathcal{Z}}_o^{\mathrm{ref}}$, that are the actual and target output spike trains of an output neuron $o$ respectively. The \ac{vRD} can then be determined from the definition \cite{Rossum2001}:
\begin{equation} \label{eq:vRD}
	\mathcal{D}(\mathcal{Z}_o, \mathcal{Z}_o^{\mathrm{ref}}) = \frac{1}{\tau_c} \int_0^{\infty} [\tilde{\mathcal{Z}}_o(t) - \tilde{\mathcal{Z}}_o^{\mathrm{ref}}(t)]^2 \mathrm{d}t \;.
\end{equation}
Using Eq. \ref{eq:vRD}, the \ac{vRD} between an output generated by the network and every potential target output is computed, giving the list of distances $\mathcal{\mathbf{D}} = \{\mathcal{D}_1,\mathcal{D}_2, ... ,\mathcal{D}_c\}$ for a total of $c$ different classes. A correct classification of the input is then made if its target class label $l$ matches the index of the minimum distance, that is if $l = \argmin_i \mathcal{\mathbf{D}}$ for $\mathcal{D}_i \in \mathcal{\mathbf{D}}$.

For a network containing more than one output neuron, responses consist of spatio-temporal output patterns $\mathcal{Z}_o \in \mathbf{Z}$ with corresponding target outputs $\mathcal{Z}_o^{\mathrm{ref}} \in \mathbf{Z}^{\mathrm{ref}}$. To compute the distance between two spatio-temporal spike patterns, the \ac{vRD} is summed over every output neuron:
\begin{equation} \label{eq:vRD_spatio}
\mathcal{D}(\mathbf{Z}, \mathbf{Z}^{\mathrm{ref}}) 
= \frac{1}{\tau_c} \sum_o \int_0^{\infty} [\tilde{\mathcal{Z}}_o(t) - \tilde{\mathcal{Z}}_o^{\mathrm{ref}}(t)]^2 \mathrm{d}t \;,
\end{equation}
and is determined with respect to each class. Similarly to a network containing a single output neuron, a correct classification of an input is made if its target class label matches the index of the minimum spatio-temporal distance.

\subsection*{Performance and convergence measures}

The classification performance of the network was taken as an exponential moving average $\tilde{\mathcal{P}}_c$ up to the $n^{\mathrm{th}}$ episode, given by $\tilde{\mathcal{P}}_c(n) = (1 - \lambda) \tilde{\mathcal{P}}_c(n-1) + \lambda \mathcal{P}_c(n)$. On each episode, the performance either took a value of $\mathcal{P}_c = \SI{100}{\%}$ for a correct input classification or otherwise $\mathcal{P}_c = 0$ (c.f. Pattern recognition). The timing parameter was taken as $\lambda = 2 / (1 + 20 p)$, which corresponded to an averaging window of $20 p$ for a total of $p$ input patterns. The \ac{vRD} was also taken as a moving average $\tilde{\mathcal{D}}$, with the same averaging window as used for $\tilde{\mathcal{P}}_c$. A moving average of each measure was necessary, given our choice of a stochastic neuron model that gave rise to fluctuating network responses between episodes.

In our simulations we measured the number of episodes taken for the network to converge in learning, that was defined in terms of its classification performance $\tilde{\mathcal{P}}_c$. Specifically, given a total of $N$ learning episodes, we considered that learning had converged on the $n^{\mathrm{th}}$ episode for the first value $\tilde{\mathcal{P}}_c(n) > 0.99\; \tilde{\mathcal{P}}_c(N)$, by which point the network performance fell within \SI{1}{\%} of its final value. In the case of the network failing to learn any input patterns, with $\tilde{\mathcal{P}}_c(N) = 0$, the number of episodes to convergence was taken as 0.

\subsection*{Simulation details}

In all simulations, we used a fixed number $n_i = 100$ of input neurons and a variable number $n_h$ of hidden neurons. Depending on the learning task, either a single output neuron or multiple output neurons determined the response of the network to presented input patterns. The simulation time step was set to $\delta t = \SI{1}{ms}$.

\paragraph{Multilayer networks.} In all simulations of a multilayer network, hidden layer synaptic weights were initialized by independently selecting each value from a uniform distribution over the range: $w_{hi} \in [0, 3)$, that gave rise to an initial hidden neuron firing rate of $\sim \SI{24}{Hz}$. Output weights were initialized depending on the number of output neurons. During learning, hidden weights were constrained to the range: $0 \leq |w_{hi}| \leq 100$, and were free to take either positive or negative values. To increase the number of eligible synapses available to the network, and to increase the diversity of hidden neuron spiking, axonal conduction delays were introduced between the input and hidden layers \cite{Bohte2004,Izhikevich2006}. Conduction delays were selected from a uniform distribution over the range $d_{hi} \in (0, 40]$ \si{ms} and rounded to the nearest \SI{1}{ms}, where $d_{hi}$ was the conduction delay between the $i^{\mathrm{th}}$ and $h^{\mathrm{th}}$ input and hidden neurons respectively. Hence, a conduction delay $d_{hi}$ resulted in a \ac{PSP} evoked at $h$ due to an input spike $t_i^f$ with an effective time course of $\epsilon(t-t_i^f-d_{hi})$ (c.f. Eq. \ref{eq:kernels}). Conduction delays were neglected between the hidden and output layers. Hidden and output layer learning rates were set to $\eta_h = 4 / (n_i\; n_o\; n_s)$ and $\eta_o = 0.02 / n_h$ respectively, where it was indicated through preliminary simulations that the dependence of $\eta_h$ on the number of output neurons $n_o$ and number of target output spikes $n_s$ dominated over the number of input patterns $p$. Both $\eta_h$ and $\eta_o$ depended on the number of afferent synapses: $n_i$ and $n_h$ respectively.

\paragraph{Single outputs.} In simulations of a multilayer network with a single output neuron, initial values of output synaptic weights were all set to the same value $w_{oh} = 12 / n_h$ that drove the output firing rate to $\sim \SI{1}{Hz}$. Each initial value of $w_{oh}$ was identical to allow equal contributions from every hidden layer neuron at the start of learning. During learning, output weights were constrained to the range $0.01 \leq w_{oh} \leq 100$; the lower bound of 0.01 was enforced to enable hidden weight updates to keep taking place, given that updates depended on output weight values according to $\Delta w_{hi} \propto w_{oh}$ (Eq. \ref{eq:w_hi_update}). Values of $w_{oh}$ were positive and prevented from changing sign during learning to ensure sufficient excitatory drive in the output neuron from the hidden layer. Preliminary simulations indicated that constraining output weights to positive values for a single output neuron had no adverse impact on learning.

\paragraph{Multiple outputs.} In simulations of a multilayer network with multiple output neurons, output synaptic weights were initialized by independently selecting each value from a uniform distribution over the range $w_{oh} \in [0, 30 / n_h)$, that drove the firing rate of each output neuron to $\sim \SI{1}{Hz}$. Randomizing output weights was necessary to increase the diversity between output responses, which improved learning in the initial stages of each simulation run.
Output weights were constrained to the range $0 \leq |w_{oh}| \leq 100$, and were allowed to change sign during learning.

\paragraph{Single-layer networks.} In simulations of a single-layer network, synaptic weights were initialized by independently selecting each value from a uniform distribution over the range $w \in [0, 1.7)$, that gave rise to an initial output firing rate of $\sim \SI{1}{Hz}$. The learning rate was set to $\eta = 4 / n_i$ and weights were constrained to the range $0 \leq |w_{oh}| \leq 100$, where the values of weights were allowed to change sign during learning. For a closer comparison, the model and parameter set used to generate output spikes in the single-layer network matched those used to generate output spikes in the multilayer network.

\section*{Acknowledgments}

BG was funded by EPSRC grant EP/J500562/1. AG and IS were funded by the Human Brain Project (HBP).

\end{document}